\documentclass[10pt,twocolumn,letterpaper]{article}

\usepackage[pagenumbers]{cvpr} 

\usepackage[utf8]{inputenc} 
\usepackage[T1]{fontenc}    

\usepackage{graphicx}
\usepackage{amsmath}
\usepackage{amssymb}

\usepackage{cite}
\usepackage{amsmath,amssymb,amsfonts}
\usepackage{algorithmic}
\usepackage{subcaption}
\usepackage{tabularx}
\usepackage{booktabs}       

\usepackage{amsmath}
\usepackage{amssymb}
\usepackage{amsmath,bm}
\usepackage{color}
\usepackage{xcolor,soul}
\usepackage{algorithm}
\usepackage{multirow}
\usepackage{xspace}
\usepackage{soul}

\usepackage{enumitem}

\usepackage{newtxtext} 

\newdimen\abovecrulesep
\newdimen\belowcrulesep
\makeatletter
\patchcmd{\@@@cmidrule}{\aboverulesep}{\abovecrulesep}{}{}
\patchcmd{\@xcmidrule}{\belowrulesep}{\belowcrulesep}{}{}

\definecolor{demphcolor}{RGB}{144, 144, 144}
\definecolor{mygray}{gray}{0.4}
\definecolor{lightgray}{rgb}{0.9, 0.9, 0.9}
\newcommand{\demph}[1]{\textcolor{demphcolor}{#1}}

\newlength\savewidth
\newcommand\shline{\noalign{\global\savewidth\arrayrulewidth\global\arrayrulewidth 1pt}\hline\noalign{\global\arrayrulewidth\savewidth}}

\newcommand{\tablestyle}[2]{\setlength{\tabcolsep}{#1}\renewcommand{\arraystretch}{#2}\centering\small}
\makeatletter\renewcommand\paragraph{\@startsection{paragraph}{4}{\z@}{.5em\@plus1ex\@minus.2ex}{-.5em}{\normalfont\normalsize\bfseries}}
\makeatother

\newcommand{\modelname}{\textbf{HiTeA}\xspace}

\setstcolor{blue}
\def\revised{\textcolor{black}}

\usepackage{hyperref}
\hypersetup{pagebackref,breaklinks,colorlinks}

\usepackage[capitalize]{cleveref}
\crefname{section}{Sec.}{Secs.}
\Crefname{section}{Section}{Sections}
\Crefname{table}{Table}{Tables}
\crefname{table}{Tab.}{Tabs.}

\begin{document}

\title{\textbf{HiTeA}: Hierarchical Temporal-Aware Video-Language Pre-training}

\author{Qinghao Ye
\quad Guohai Xu
\quad Ming Yan\thanks{Corresponding Author.}
\quad Haiyang Xu \\
Qi Qian
\quad Ji Zhang
\quad Fei Huang \\ \\
DAMO Academy, Alibaba Group \\
}
\maketitle

\begin{abstract}
  Video-language pre-training has advanced the performance of various downstream video-language tasks. However, most previous methods directly inherit or adapt typical image-language pre-training paradigms to video-language pre-training, thus not fully exploiting the unique characteristic of video, i.e., temporal. In this paper, we propose a \textbf{Hi}erarchical \textbf{Te}mporal-\textbf{A}ware video-language pre-training framework, \modelname, with two novel pre-training tasks for 
 modeling cross-modal alignment between moments and texts as well as the temporal relations of video-text pairs. Specifically, we propose a cross-modal moment exploration task to explore moments in videos, which results in detailed video moment representation. 
Besides, the inherent temporal relations are captured by aligning video-text pairs as a whole in different time resolutions with multi-modal temporal relation exploration task. Furthermore, we introduce the shuffling test to evaluate the temporal reliance of datasets and video-language pre-training models. We achieve state-of-the-art results on 15 well-established video-language understanding and generation tasks, especially on temporal-oriented datasets (\eg, SSv2-Template and SSv2-Label) with 8.6\% and 11.1\% improvement respectively. \modelname also demonstrates strong generalization ability when directly transferred to downstream tasks in a zero-shot manner. Models and demo will be available on \href{https://www.modelscope.cn/}{ModelScope}.
  
\end{abstract}

\vspace{-2ex}
\section{Introduction}
Vision and language are two primary signals that constitute the real-world perception of humanity. With the success of image-language pre-training \cite{chen2020uniter,li2021align, wang2021simvlm,li2022mplug}, video-language pre-training \cite{miech2020milnce, li2020hero, li2022alpro, li2022lavender} has recently received increasing attention. 
Large-scale video-language pre-training helps the model to learn effective multi-modal representation, which has shown significant improvement on a variety of video-language downstream tasks, such as video-text retrieval, video question answering and video captioning \cite{xu2017msrvttqa, xu2016msrvtt, chen2011msvd, xiao2021nextqa, yu2019activitynetqa, li2021sumgda, maharaj2017lsmdcfib, torabi2016lsmdcmc}.

\begin{figure}
    \centering
    \includegraphics[width=0.99\linewidth]{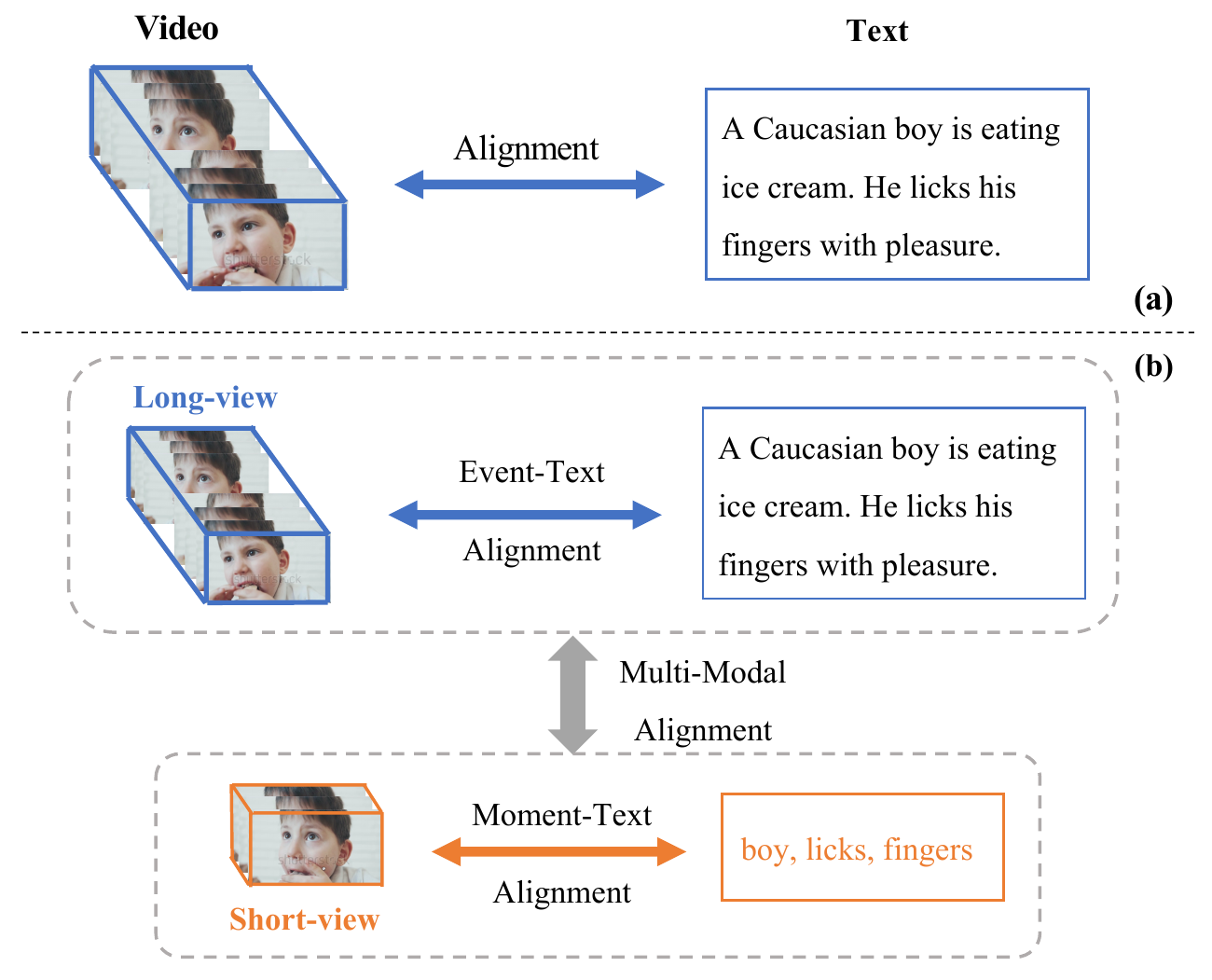}
\caption{Comparison between existing paradigms and ours for video-language pre-training. (a) Previous methods align video and text within global perspective as the pretext. (b) We introduce \modelname by varying video in different temporal views and modeling cross-modal alignment between moments and texts, as well as the temporal relations between multi-modal pairs.}
\label{fig:1}
\vspace{-2ex}
\end{figure}

Inspired by the success of image-language pre-training paradigm, various methods \cite{li2022lavender, li2022alpro, fu2021violet, ge2022bridgeformer, li2020hero} have been proposed to adapt it to video-language pre-training. ClipBERT \cite{lei2021clipbert} and Singularity \cite{lei2022singularity} directly build on representations from image encoders and aggregate them via score aggregation function and temporal encoder. Furthermore, MIL-NCE \cite{miech2020milnce} and Frozen \cite{bain2021frozen} switch image encoder to video encoder for spatio-temporal video representation learning and align the video with corresponding text. In addition, some advanced pre-training tasks are designed through modeling entity \cite{li2022alpro}, reconstructing masked patches \cite{fu2021violet} and predicting frame order \cite{li2020hero, zellers2021merlot}. Despite their promising performance on downstream tasks, they treat video within global perspective illustrated in Figure \ref{fig:1}(a), thus failing to consider fine-grained temporal information and relations which are essential to video-language pre-training.

Since untrimmed video contains various temporal details, directly treating the video globally has two main limitations: (1) Less effective in modeling the fine-grained moment information including atomic actions and moments. As illustrated in Figure \ref{fig:1}(b), we vary time resolutions and generate two views (long \& short) for the input video. As a result, the shot-view video clip tends to represent the moment information and the long-view video may express more event-level information. For example, the short-view video clip in Figure \ref{fig:1}(b) only describes the moment of "lick fingers" rather than "eating ice cream". Such fine-grained moment information is hard to be captured by the long-view video under global event perspective; (2) Ignoring the temporal relations implicitly existed in the video. Knowing the event expressed by the text, the moment "eating ice cream" can be inferred from the moment "lick fingers" shown by short-view video. However, the implicit temporal relations between the moments are rarely explored in previous works.

To address these problems, we propose a  \textbf{Hi}erarchical \textbf{Te}mporal-\textbf{A}ware video-language pre-training framework, \modelname, for both multi-modal understanding and generation. 
Except for the standard pre-training tasks, \modelname introduces two novel temporal-aware video-language pre-training tasks, named \textit{cross-modal moment exploration} (CME) and \textit{multi-modal temporal relation exploration} (MTRE), 
which not only model the fine-grained moments with partial cross-modal alignment but also capture temporal relations between multi-modal pairs hierarchically.
Specifically, we first generate the long-view and short-view videos with different time resolutions to build hierarchy of the input video. Then, based on the similarities of words and short-view video, we select the most relevant words as positive and leave the rest of the words as hard negatives. The CME pre-training task is applied to align the positive words and shot-view video representations in the same embedding space.
Moreover, to capture association between moments and the event, we match different views for the same video. However, directly matching two views visually would be noisy due to the background similarity \cite{roth2022languagemetric}. 
To this end, we perform multi-modal alignment between video-text pairs via the MTRE pre-training task.
More specifically, the shot-view video guided by most relevant words and the long-view video guided by text will be aligned. 
Empowered by above two novel temporal-aware video-language pre-training tasks, \modelname captures both fine-grained moment information and temporal relations between different views of video.
 
In spite of a good performance, recent studies \cite{lei2022singularity, buch2022atp} reveal most video-language downstream datasets are biased towards still objects, scenes, \etc., while the temporal dynamics are negligible. To evaluate the temporal performance of the video-language pre-training model and temporal reliance of downstream datasets, we introduce temporal shuffling test for these datasets. This enables a more comprehensive evaluation of temporal modeling capability in the video-language pre-training field. Besides, our method achieves significant improvement on the datasets with heavy temporal reliance.

In summary, our key contributions are the followings:
\begin{itemize}\setlength{\itemsep}{-1pt}
    \item We propose a novel hierarchical temporal-aware video-language pre-training framework with both video-language understanding and generation capabilities.
    \item We introduce additional temporal-aware pre-training tasks by performing cross-modal and multi-modal alignment hierarchically, which not only model moment information with fine-grained semantics but also capture temporal relations between moments and event.
    \item Extensive experiments demonstrate the effectiveness of \modelname, and it achieves state-of-the-art performance on 15 video-language downstream datasets including video-text retrieval, video question answering, and video captioning, especially on temporal-oriented datasets (\eg, SSv2-Template and SSv2-Label) with 8.6\% and 11.1\% improvement respectively.
\end{itemize}

\begin{figure*}
    \centering
    \includegraphics[width=0.85\textwidth]{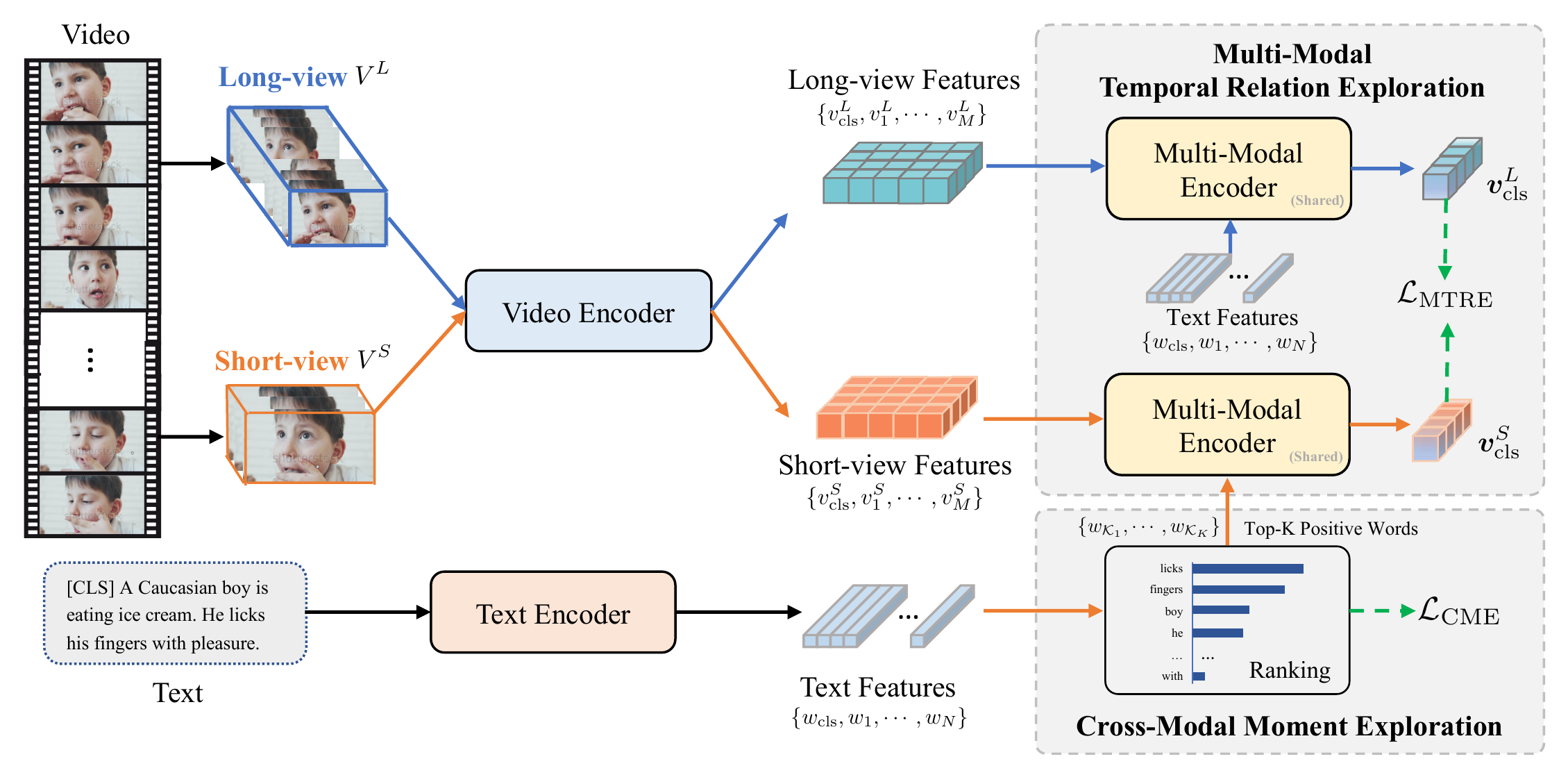}
    \caption{Illustration of the proposed \modelname. We first generate two different temporal views for the input video, where the long-view is the video itself and the short-view is randomly truncated from the input video. To explore the moment revealed in the short-view, \textit{cross-modal moment exploration} (CME) selects the candidate words from the input text with $\mathcal{L}_{\text{CME}}$. Then, we perform \textit{multi-modal temporal relation exploration} (MTRE) for modeling the temporal relations between two video-text pairs with different views by $\mathcal{L}_{\text{MTRE}}$. Note that the multi-modal encoders and the text features are shared.}
    \vspace{-3ex}
\label{fig:2}
\end{figure*}


\section{Related Work}

\paragraph{Video-Language Pre-training}

Benefiting from a large number of image/video-text pairs, video-language pre-training (VLP) exhibits superior capabilities on various video-text benchmarks. The method of VLP is constantly evolving. Traditional approaches \cite{sun2019videobert, zhu2020actbert, luo2020univl, li2020hero} leverage offline-extracted dense video features for pre-training to circumvent the expensive computation overhead. In contrast, ClipBERT \cite{lei2021clipbert} suggests that sparse sampling can enable affordable end-to-end learning and improve performance simultaneously. Recent emerging approaches \cite{bain2021frozen, li2022alpro, ge2022bridgeformer, lei2021clipbert, huang2022clover, li2022lavender} adopt this strategy and propose new model architectures and pre-training tasks. Frozen \cite{bain2021frozen} trains jointly on image and video datasets via video-text contrastive learning (VTC). ALPRO \cite{li2022alpro} proposes a new visually-grounded pre-training task combined with VTC, video-text matching (VTM) and masked language modeling (MLM) \cite{devlin2018bert} to learn fine-grained region-entity alignment. LAVENDER \cite{li2022lavender} formulates all pre-training and downstream tasks as MLM so that a unified architecture can used for all video-text tasks. Apart from above representative works, frame order modeling (FOM) \cite{li2020hero, zellers2021merlot} and masked video modeling (MVM) \cite{fu2021violet} are designed for VLP. 
However, the temporal characteristic of video still remains largely unexplored. To this end, we introduce a novel hierarchical temporal-aware VLP framework which not only models the fine-grained moment information but also captures their correlations with different temporal granularities.

\paragraph{Temporal Modeling}

The temporal characteristic acts as a vital role in VLP since it provides the model with the capabilities of reasoning and understanding causality. Previous efforts in this field can be roughly divided into three categories. First, several methods directly transfer image-text models to video-text tasks by simply concatenating video frame \cite{li2022blip,li2022mplug} or building a additional temporal encoder \cite{luo2022clip4clip,ma2022xclip}. Second, some works \cite{bain2021frozen,li2022alpro,fu2021violet,li2022lavender} switch the image encoder to video encoder for learning spatio-temporal contexts within videos. Third, HERO \cite{li2020hero} and MERLOT \cite{zellers2021merlot} design FOM task to explicitly recover the correct temporal order of shuffled frames. Nonetheless, ATP \cite{buch2022atp} and Singularity \cite{lei2022singularity} reveal the existence of a static appearance bias in popular video-language datasets, \revised{and} they develop single-frame models to achieve surprisingly strong performance, comparable or even better than above methods with explicit temporal modeling. Therefore, they recommend SSv2 \cite{lei2022singularity} and NExT-QA \cite{xiao2021nextqa} datasets to test the temporal ability of VLP models. 
\revised{Different from previous approaches, we vary the temporal resolutions and generate two views of video so as to construct the temporal hierarchy, which equips the model with the ability to learn both fine-grained moment information and temporal relations at the same time.}

\section{Method}

\subsection{Overview}
\label{sec:overview}

Figure~\ref{fig:2} sketches the overview of the \modelname. In concrete, our model consists of two unimodal encoders for encoding video and text separately, a multi-modal encoder for video and text interaction, and a text decoder for generation which is omitted here for simplicity and detailed in Appendix. 


For video representation, previous methods \cite{lei2021clipbert,li2022alpro,li2022lavender} encode the whole input video as a single-view feature, ignoring the rich temporal details contained in the video. Thus, we first treat the video into two views with different time resolutions to build hierarchy of the input video. Specially, the untrimmed video is regarded as a long-view video $V^L$ for capturing event information, and a video segment is randomly truncated from the input video as the short-view for capturing moment information denoted as $V^S$. Then, we use the video encoder to encode an arbitrary view of video $V \in \mathbb{R}^{T\times H \times W}$ into a sequence of embeddings: $\mathcal{V} = \{v_{\text{cls}}, v_1, \cdots, v_M\} \in \mathbb{R}^{M\times D}$, where $M$ is the number of flattened patches, $D$ is hidden size, and $v_{\text{cls}}$ is the embedding of the visual [CLS] token which provides global representation of the video. For text representation, we use the text encoder to transform the text $\textit{T}$ into a sequence of embeddings: $\mathcal{T} = \{w_{\text{cls}}, w_1, \cdots, w_N\} \in \mathbb{R}^{N\times D}$, where $N$ is length of the text. After that, the multi-modal encoder takes video features $\mathcal{V}$ and text features $\mathcal{T}$ as inputs and yields the multi-modal representation $\bm{v}_{\text{cls}}$ for the video.

In order to take full advantage of the different view of the video, we introduce \textit{cross-modal moment exploration} (CEM) to explore the proper words or phrases from input text to align the short-view video with $\mathcal{L}_{\text{CME}}$ for capturing the moment information in Section~\ref{sec:cme}. Furthermore, to model the relations between the short-view video containing moment information and the long-view video with event information, we propose \textit{multi-modal temporal relation exploration} (MTRE) 
to align the multi-modal representation of short-view and long-view videos by $\mathcal{L}_{\text{MTRE}}$ in Section~\ref{sec:mtre}. Lastly, we introduce the overall pre-training objective for training the model in Section \ref{sec:loss}.



\subsection{Cross-Modal Moment Exploration}
\label{sec:cme}

To understand the fine-grained moment information, the video with short temporal range (\ie short-view of video) should be aligned with the corresponding text. However, since the video is partially aligned with the paired text which describes the whole video, directly aligning short-view of video with paired text would bring noise to model learning and degrade the performance. Therefore, we propose a novel pre-training task named \textit{cross-modal moment exploration} (CME), which enables the model to understand fine-grained moment information.

Formally, we first discover the possible positive words for the video in short-view by computing the cosine similarity of the word embedding sequence $\{w_1, \cdots, w_N\}$ from text encoder and the short-view video representation $v_{\text{cls}}^S$ from video encoder as:
\begin{align}
    \mathcal{K} = \{\pi(1), \cdots, \pi(K)\},
\label{eq:topk}
\end{align}
where $\pi:\{1, \cdots, N\}\rightarrow \{1, \cdots, N\}$ is a permutation function for ranking such that $s(w_{\pi(1)}, v_{\text{cls}}^S) \geq \cdots \geq s(w_{\pi(N)}, v_{\text{cls}}^S)$, and $\mathcal{K}$ is the set of selected word indices, $K$ is the number of possible selected words, and $s(x, y) = x^Ty / \|x\|_2 \|y\|_2$ represents the cosine similarity between $x$ and $y$. After obtaining the words for the video in short-view as the positive pair, the cross-modal moment exploration loss $\mathcal{L}_{\text{CME}}$ is computed with negative pairs from other words in the input text, which is defined as:
\begin{align}
    \mathcal{L}_{\text{CME}} = -\frac{1}{B}  \sum_{i=1}^B 
\left( 
\frac{1}{|\mathcal{K}|} \sum_{k \in \mathcal{K}} \log
\frac{\exp((v_{\text{cls}}^S)^\top_i w_{i,k}/\tau)} {\sum_{n=1}^{N} \exp((v_{\text{cls}}^S)^\top_i w_{i,n}/\tau)}
\right),
\label{eq:fine_grained_loss}
\end{align}
where $\tau$ is the learnable temperature hyper-parameter that controls the sharpness of the output distribution, and it is initialized as 0.07. As a consequence, the model is able to understand moment information via the proposed cross-modal exploration scheme.

\begin{table*}[t!]
\centering
    \tablestyle{7pt}{1.1} 
    \def \w{15pt}
    \resizebox{0.9\linewidth}{!}{
    \begin{tabular}{ll|ccc|ccc|ccc|ccc}
        \shline
        ~ & ~ & \multicolumn{3}{c}{MSRVTT}& \multicolumn{3}{c}{DiDeMo} & \multicolumn{3}{c}{LSMDC} & \multicolumn{3}{c}{ActivityNet Caption}\\
        \cmidrule(lr){3-5} \cmidrule(lr){6-8} \cmidrule(lr){9-11} \cmidrule(lr){12-14}
        Method & \# PT Data & R@1 & R@5 & R@10 & R@1 & R@5 & R@10 & R@1 & R@5 & R@10 & R@1 & R@5 & R@10 \\
        \shline
        ClipBERT~\cite{lei2021clipbert} & 0.2M & 22.0 & 46.8 & 59.9 & 20.4 & 48.0 & 60.8  & - & - & - & 21.3 & 49.0 & 63.5 \\
        Frozen~\cite{bain2021frozen} & 5M  & 31.0 & 59.5 & 70.5 & 31.0 & 59.8 & 72.4 & 15.0 & 30.8 & 39.8 & - & - & - \\
        ALPRO~\cite{li2022alpro} & 5M & 33.9 & 60.7 & 73.2 & 35.9 & 67.5 & 78.8   & - & - & - & - & - & - \\
        BridgeFormer~\cite{ge2022bridgeformer} & 5M  & 37.6 & 64.8 & 75.1 & 37.0 & 62.2 & 73.9  & 17.9 & 35.4 & 44.5 & - & - & - \\
        Singularity~\cite{lei2022singularity} & 5M & 36.8 & 65.9 & 75.5 & 47.4 & 75.2 & 84.0 & - & - & - & 43.0 & 70.6 & 81.3 \\
        LAVENDER~\cite{li2022lavender} & 5M & 37.8 & 63.8 & 75.0 & 47.4 & 74.7 & 82.4 & 22.2 & 43.8 & 53.5 & - & - & - \\
         \hline
         \multicolumn{5}{l}{\demph{\textit{Models pre-trained on more data}}}\\
         \hline
        \demph{VIOLET~\cite{fu2021violet}} & \demph{183M}  & \demph{34.5} & \demph{63.0} & \demph{73.4} & \demph{32.6} & \demph{62.8} & \demph{74.7}  & \demph{16.1} & \demph{36.6} & \demph{41.2} & \demph{-} & \demph{-} & \demph{-} \\
        \demph{All-in-one~\cite{wang2022allinone}} & \demph{138M}  & \demph{37.9} & \demph{68.1} & \demph{77.1} & \demph{32.7} & \demph{61.4} & \demph{73.5}  & \demph{-} & \demph{-} & \demph{-} & \demph{22.4} & \demph{53.7} & \demph{67.7} \\
        \demph{Clip4Clip~\cite{luo2022clip4clip}} & \demph{400M} &  \demph{42.1} & \demph{71.9} & \demph{81.4} & \demph{43.4} & \demph{70.2} & \demph{80.6}  & \demph{21.6} & \demph{41.8} & \demph{49.8} & \demph{40.5} & \demph{72.4} & \demph{-} \\
        \demph{X-CLIP~\cite{ma2022xclip}} & \demph{400M} &  \demph{46.1} & \demph{73.0} & \demph{83.1} & \demph{45.2} & \demph{74.0} & \demph{-}  & \demph{23.3} & \demph{43.0} & \demph{-} & \demph{44.3} & \demph{74.1} & \demph{-} \\
        \hline
        \modelname & 5M & \textbf{44.4} & \textbf{69.3} & \textbf{78.9} & \textbf{51.8} & \textbf{79.1} & \textbf{85.3} &  \textbf{27.1} & \textbf{46.2} & \textbf{54.5} & \textbf{45.1} & \textbf{73.5} & \textbf{84.2} \\
        \demph{\modelname} & \demph{17M} & \demph{\textbf{46.8}} & \demph{\textbf{71.2}} & \demph{\textbf{81.9}} & \demph{\textbf{56.5}} & \demph{\textbf{81.7}} & \demph{\textbf{89.7}} &  \demph{\textbf{28.7}} & \demph{\textbf{50.3}} & \demph{\textbf{59.0}} & \demph{\textbf{49.7}} & \demph{\textbf{77.1}} & \demph{\textbf{86.7}} \\
        \shline
    \end{tabular}
    }
    \vspace{-1ex}
    \caption{\textbf{Performance comparison on text-to-video retrieval.} All results are reported on R@1/R@5/R@10. We gray out methods that use significantly more pre-training data for fair comparison. \textbf{\# PT Data} is the number of video-text pairs for pre-training. 
    }
    \label{table:retrieval-sota}
    \vspace{-2ex}
\end{table*}

\subsection{Multi-Modal Temporal Relation Exploration}
\label{sec:mtre}
While the video encoder has demonstrated its effectiveness in learning temporal representation implicitly \cite{fu2021violet, li2022alpro, ye2021temporal}, it remains a challenge to discover the inherent temporal relations. As a result, the limited capabilities in temporal modeling deteriorate the downstream task in temporal reasoning. This is in particular a missing point for the existing video-language pre-training paradigm \cite{lei2021clipbert,li2022alpro, ge2022bridgeformer, li2022lavender}, which usually focuses on bridging video and text neglecting the function of text for guiding the video context representation learning thus losing the temporal cues. 

To this end, we introduce \textit{multi-modal temporal relation exploration} (MTRE), a novel temporal-aware pre-training task that improves models' capacities in capturing temporal correlation of moments in video with fine-grained text guidance by aligning multi-modal pairs. Specially, the short-view video $V^S$ would represent moment information with respect to the whole video. On the contrary, the long-view video $V^L$ 
expresses the event and topical information. To obtain the text-guided video features, we feed videos in different temporal views into the video encoder individually. Then, the text features are extracted and interact with the video features by the multi-modal encoder and yield text-guided video representations $\bm{v}_{\text{cls}}^L \in \mathbb{R}^D$ and $\bm{v}_{\text{cls}}^S \in \mathbb{R}^D$ as follows:
\begin{align}
    \bm{v}_{\text{cls}}^L &= f(\{v_{\text{cls}}^L, v_1^L, \cdots, v_M^L\},\{w_{\text{cls}}, w_1, \cdots, w_N\}), \\
    \bm{v}_{\text{cls}}^S &= f(\{v_{\text{cls}}^S, v_1^S, \cdots, v_M^S\}, \{w_{\text{cls}}, w_1, \cdots, w_N\}),
\label{eq:fusing}
\end{align}
where $f(\mathcal{V}, \mathcal{T})$ represents the multi-modal encoder with video features $\mathcal{V}$ and text features $\mathcal{T}$.
However, since the short-view of the video is partially aligned with the text, using the whole text is not reasonable for generating accurate text-guided video feature for short-view. 
Meanwhile, improper video-text pairs would yield noisy multi-modal representation thus degrading the performance of the model. 
Therefore, thanks to the positive words mined by \textit{cross-modal moment exploration}, we can calibrate representation for short-view video by:
\begin{align}
    \bm{v}_{\text{cls}}^S &= f(\{v_{\text{cls}}^S, v_1^S, \cdots, v_M^S\}, \{w_{\mathcal{K}_1}, \cdots, w_{\mathcal{K}_K}\}),
\end{align}
where $\mathcal{K}_i \in \mathcal{K}$ is the index of the set for selected positive words. Then, we aim to match the representation of produced text-guided video features in different granularities in order to enable the model to predict the past and the future from the short-view of video, which benefits for capturing the general structure of the video. Specifically, we adopt the SimSiam framework \cite{chen2021simsiam} for minimizing their negative cosine similarity:
\begin{align}
    \mathcal{D}(p^S, z^L) = -\frac{p^S}{\|p^S\|_2} \cdot \frac{z^L}{\|z^L\|_2},
\end{align}
where $p^S = h(g(\bm{v}_{\text{cls}}^S))$ and $z^L = g(\bm{v}_{\text{cls}}^L)$. The $g$ and $h$ are projection MLP head and prediction MLP head \cite{grill2020byol, chen2021moco}. Minimizing $\mathcal{D}(p^S, z^L)$ is equivalent for minimizing the mean square error between $p^S$ and $z^L$, which encourages the videos in different temporal magnitudes to be similar. Following \cite{chen2021simsiam, grill2020byol}, we defined a symmetrized loss as:
\begin{align}
    \mathcal{L}_{\text{MTRE}} = \frac{1}{2}\left[ \mathcal{D}(p^L, \texttt{sg}(z^S)) + \mathcal{D}(p^S, \texttt{sg}(z^L))  \right],
\label{eq:loss_slm}
\end{align}
where $\texttt{sg}(\cdot)$ is the stop-gradient operation that prevents the model from collapse during training \cite{chen2021simsiam}. 

\subsection{Pre-training Objectives}
\label{sec:loss}
Apart from the two proposed temporal-aware pre-training tasks, we follow proven video-text pre-training approaches \cite{li2022alpro, bain2021frozen, li2022lavender} to adopt the standard pre-training tasks including video-text contrastive (VTC), video-text matching (VTM), masked language modeling (MLM), and prefix language modeling (PrefixLM) described in the related work. Precisely, VTC and VTM align the video and text from the global perspective, while MLM and PrefixLM contribute to multi-modal understanding and generation capabilities of the model. Details of these objectives are described in the Appendix. We simply combine these as the base training objective $\mathcal{L}_{base}$ for our model. Therefore, the full pre-training objective is computed as:
\begin{equation}
    \mathcal{L} = \mathcal{L}_{\text{base}} + \mathcal{L}_{\text{CME}} + \mathcal{L}_{\text{MTRE}}.
\end{equation}

\section{Experiments}
\subsection{Experiment Setup}
\paragraph{Pre-training Datasets}
Following the recent work \cite{bain2021frozen, li2022alpro, lei2022singularity, ge2022bridgeformer, li2022lavender}, we pre-train our model on a webly-sourced video dataset WebVid-2M \cite{bain2021frozen} with 2.5M video-text pairs and a image-text dataset Google Conceptual Captions (CC3M) \cite{sharma2018conceptual} with 3M image-text pairs. Unlike previous methods, we do not pre-train our model on the large-scale video-text datasets like HowTo100M \cite{miech2020milnce} with 136M video-text pairs and YT-Temporal-180M \cite{zellers2021merlot} due to the heavy computation. For scaling up, we also trained our model on the widely used image-text pre-training datasets including MS COCO \cite{lin2014mscoco}, Visual Genome \cite{krishna2017vg}, SBU Captions \cite{ordonez2011sbu} and Conceptual 12M \cite{sharma2018conceptual}, we refer this setting as 17M corpus.

\paragraph{Downstream Datasets}
We evaluate our pre-trained model on 18 video-language benchmarks including video-text retrieval, video question answering, and video captioning tasks. Specifically, video question answering (VideoQA) can be categorized as Multiple-Choice (MC) and Open-Ended (OE) settings. The evaluation datasets are briefly summarized in below. The details can be found in the Appendix.
\begin{itemize}[leftmargin=*]\setlength{\itemsep}{-2pt}
    \item \textbf{Video-Text Retrieval}: MSRVTT \cite{xu2016msrvtt}, DiDeMo \cite{anne2017didemo}, LSMDC \cite{rohrbach2015lsmdc}, ActivityNet Caption \cite{krishna2017activitynetret}, SSv2-Label \cite{lei2022singularity}, and SSv2-Template \cite{lei2022singularity};
    \item \textbf{VideoQA (MC)}: TGIF-Action, TGIF-Transition \cite{jang2017tgif}, MSRVTT-MC \cite{yu2018msrvttmc}, LSMDC-MC \cite{torabi2016lsmdcmc}, and NExT-QA \cite{xiao2021nextqa};
    \item \textbf{VideoQA (OE)}: TGIF-Frame \cite{jang2017tgif}, MSRVTT-QA, MSVD-QA \cite{xu2017msrvttqa}, LSMDC-FiB \cite{maharaj2017lsmdcfib} and ActivityNet-QA \cite{yu2019activitynetqa}.
    \item \textbf{Video Captioning}: MSRVTT \cite{xu2016msrvtt} and MSVD \cite{chen2011msvd}.
\end{itemize}

\paragraph{Implementation Details}
Our implementation of \modelname is based on PyTorch \cite{paszke2019pytorch}. In detail, we instantiate the video encoder with MViT-Base model \cite{li2022mvitv2} pretrained on ImageNet-21K \cite{ridnik2021imagenet}. The text encoder is initialized from first six layers of pre-trained BERT-Base \cite{devlin2018bert}, and the multi-modal encoder is initialized with last six layers of pre-trained BERT-Base. We pre-train \modelname for 10 epochs, using a batch size of 16 on 8 NVIDIA A100 GPUs. We use AdamW \cite{kingma2014adam} optimizer with a weight decay of 0.02 and betas (0.9, 0.98). The learning rate is first warmed up to 5e-5 in the first 1000 iterations, and decays following a cosine schedule. During pre-training, we sparsely sample 8 frames for short and long view while preserving their order in-between and resize them to 224 $\times$ 224. The duration of short view is restricted as the 1/8 of the whole video duration. $K$ is empirically set to 5. The MLM mask ratio is set to $15\%$. Details of fine-tuning stage are described in Appendix.

\begin{table*}[t]
\centering
    \tablestyle{4pt}{1.1} 
    \def \w{20pt} 
    \resizebox{0.75\linewidth}{!}{
    \begin{tabular}{ll|ccc|cc|cc|c|c}
        \shline
        ~ & ~ & \multicolumn{3}{c}{TGIF}  & \multicolumn{2}{c}{MSRVTT}  & \multicolumn{2}{c}{LSMDC}  & \multicolumn{1}{c}{MSVD} & \multicolumn{1}{c}{ActivityNet}\\
        \cmidrule(lr){3-5} \cmidrule(lr){6-7} \cmidrule(lr){8-9} \cmidrule(lr){10-10} \cmidrule(lr){11-11}
        Method & \#PT Data & Action & Transition & Frame  & MC & QA  & MC & FiB  & QA & QA \\
        \shline
        ClipBERT~\cite{lei2021clipbert} & 0.2M & 82.8 & 87.8 & 60.3  & 88.2 & 37.4  & - & -  & -  & -  \\
        ALPRO~\cite{li2022alpro} & 5M & - & - & - & - & 42.1 & - & - & 46.3 & -  \\
        Singularity~\cite{lei2022singularity} & 5M & - & - & - & 92.0 & 42.7 & - & - & - & 41.8 \\
        LAVENDER~\cite{li2022lavender} & 5M & 96.6 & \textbf{99.1} & 72.2 & 96.6 & 44.2 & \textbf{86.0} & \textbf{56.9} & 55.4 & -  \\
        Clover~\cite{huang2022clover} & 5M & 94.9 & 98.0 & 71.4 & 95.0 & 43.9 & 83.2 & 54.1 & 51.9 & - \\
         \hline
         \multicolumn{5}{l}{\demph{\textit{Models pre-trained on more data}}}\\
         \hline
        \demph{VIOLET~\cite{fu2021violet}}& \demph{183M}  & \demph{92.5} & \demph{95.7} & \demph{68.9}  & \demph{91.9} & \demph{43.9}  & \demph{82.8} & \demph{53.7}  & \demph{47.9}  & \demph{38.9} \\
        \demph{JustAsk~\cite{yang2021justask}}& \demph{69M}  & \demph{-} & \demph{-} & \demph{-}  & \demph{-} & \demph{41.5}  & \demph{-} & \demph{-}  & \demph{46.3}  & \demph{38.9} \\
        \demph{MERLOT~\cite{zellers2021merlot}} & \demph{180M}  &   \demph{94.0} & \demph{96.2} & \demph{69.5}  & \demph{90.9} & \demph{43.1}  & \demph{81.7} & \demph{52.9}  & \demph{-}  & \demph{41.4}  \\
        \demph{All-in-one~\cite{wang2022allinone}} & \demph{283M} & \demph{95.5} & \demph{94.7} & \demph{66.3}  & \demph{92.3} &  \demph{46.8} & \demph{84.4} & \demph{-}  & \demph{48.3}  & \demph{-} \\
         \hline
          \modelname & 5M & \textbf{96.8} & 98.8 & \textbf{72.5}  & \textbf{97.2} & \textbf{45.4}  & 85.8 & 54.6 & \textbf{55.6}  &  \textbf{45.1} \\
           \demph{\modelname} & \demph{17M} & \demph{\textbf{97.2}} & \demph{\textbf{98.8}} & \demph{\textbf{73.2}}  & \demph{\textbf{97.4}} & \demph{\textbf{45.9}}  & \demph{85.3} & \demph{54.5} & \demph{55.3}  &  \demph{\textbf{46.4}} \\
         \shline
    \end{tabular}
    }
    \vspace{-2ex}
    \caption{\textbf{Performance comparison on video question answering.} Accuracy is reported for evaluation. We gray out methods that use significantly more pre-training data for fair comparison. 
    }
    \label{table:videoqa-sota}
    \vspace{-1ex}
\end{table*}

\begin{table}[t!]
\centering
    \tablestyle{7pt}{1.1} 
    \def \w{15pt}
    \resizebox{0.9\linewidth}{!}{
    \begin{tabular}{ll|cc}
        \shline
        Method & \# PT Data & MSRVTT  & MSVD \\
        \hline
        UniVL \cite{luo2020univl} & 180M & 49.9 & - \\
        SwinBERT \cite{lin2022swinbert} & - & 53.8 & 120.6 \\
        MV-GPT \cite{seo2022mvgpt} & 53M & 60.0 & - \\
        CLIP4Caption \cite{tang2021clip4caption} & 400M & 57.7 & - \\
        LAVENDER \cite{li2022lavender} & 5M & 58.0 & 142.9 \\
        \hline
        \modelname & 5M & \textbf{62.5} & \textbf{145.1} \\
        \demph{\modelname} & \demph{17M} & \demph{\textbf{65.1}} & \demph{\textbf{146.9}} \\
        \shline
    \end{tabular}
    }
    \vspace{-1ex}
    \caption{\textbf{Performance comparison on video captioning.} CIDEr \cite{vedantam2015cider} is reported for evaluation.}
    \label{table:caption-sota}
    \vspace{-3ex}
\end{table}


\begin{table*}[t]
\centering
    \tablestyle{5pt}{1.1} 
    \def \w{20pt} 
    \resizebox{0.9\linewidth}{!}{
    \begin{tabular}{l|cccc cccc cc c}
        \shline
        \multirow{2}{*}{Method} & \multicolumn{4}{c}{MSRVTT-Retrieval} & \multicolumn{4}{c}{SSv2 Template-Retrieval} & \multicolumn{2}{c}{NExT-QA (Hard)} & MSVD-QA \\
        \cmidrule(lr){2-5} \cmidrule(lr){6-9} \cmidrule(lr){10-11} \cmidrule(lr){12-12}
          & R@1 & R@5 & R@10 & AveR & R@1 & R@5 & R@10 & AveR & Acc@C & Acc@T & Acc. \\
        \hline
        $\mathcal{L}_{\text{base}}$  &  40.0  & 68.0 & 77.1 & 61.7 & 80.5 & 100.0 & 100.0 & 93.5 & 44.0 & 46.4 & 52.7 \\ \hline
        $\mathcal{L}_{\text{base}}$ + $\mathcal{L}_{\text{MTRE}}$ & 41.6  & 69.1 & 78.2 & 63.0 & 83.3 & 98.9 & 100.0 & 94.1 & 46.3 & 46.4 & 54.8 \\ \hline
        $\mathcal{L}_{\text{base}}$ + $\mathcal{L}_{\text{CME}}$ &  42.0 & 69.3 & \textbf{79.7} & 63.7 & 83.9 & 99.4 & 100.0 & 94.4 & 46.3 & 48.3 & 54.3 \\ \hline
        $\mathcal{L}_{\text{base}}$ + $\mathcal{L}_{\text{CME}}$ + $\mathcal{L}_{\text{MTRE}}$ & \textbf{44.4} & \textbf{69.3} & 78.9 & \textbf{64.2} & \textbf{85.6} & \textbf{100.0} & \textbf{100.0} & \textbf{95.2} & \textbf{47.8} & \textbf{48.6} & \textbf{55.6} \\ \hline
        \shline
    \end{tabular}
    }
    \vspace{-2ex}
    \caption{Evaluation of the proposed methods on four downstream video-language tasks. For text-to-video retrieval, R@1, R@5, R@10, and the average are reported. For video question answering, we report the accuracy.}
    \label{table:ablation_loss}
    \vspace{-2ex}
\end{table*}
\begin{table}[t!]
\centering
    \tablestyle{7pt}{1.1} 
    \def \w{15pt}
    \resizebox{\linewidth}{!}{
    \begin{tabular}{ll|ccc|ccc}
        \shline
        ~ & ~ & \multicolumn{3}{c}{SSv2-Label}& \multicolumn{3}{c}{SSv2-Template} \\
        \cmidrule(lr){3-5} \cmidrule(lr){6-8} 
        Method & \# PT Data & R@1 & R@5 & R@10 & R@1 & R@5 & R@10 \\
        \hline
        Frozen~\cite{bain2021frozen} & 5M  & - & - & - & 52.9 & 94.8 & 99.4 \\
        \demph{Clip4Clip~\cite{luo2022clip4clip}} & \demph{400M} &  \demph{43.1} & \demph{71.4} & \demph{80.7} &  \demph{77.0} & \demph{96.6} & \demph{98.3} \\
        Singularity~\cite{lei2022singularity} & 5M & 44.1 & 73.5 & 82.2 & 77.0 & 98.9 & 99.4  \\
        \hline
        \modelname & 5M & \textbf{55.2} & \textbf{81.4} & \textbf{89.1} & \textbf{85.6} & \textbf{100.0} & \textbf{100.0}   \\
        \shline
    \end{tabular}
    }
    \vspace{-2ex}
    \caption{Comparison of existing methods on Something-to-Something (SSv2) text-to-video retrieval.}
    \label{table:retrieval-ssv2}
    \vspace{-2ex}
\end{table}

\begin{table}[t!]
\centering
    \tablestyle{7pt}{1.1} 
    \def \w{15pt}
    \resizebox{\linewidth}{!}{
    \begin{tabular}{ll|cccc}
        \shline
        Method & \# PT Data & Acc@C & Acc@T & Acc@D & Acc. \\
        \shline
        \textit{Full Set} \\
        \hline 
        \demph{Human} & \demph{-} & \demph{87.6} & \demph{88.6} & \demph{90.4} & \demph{88.4} \\
        \hline
        HCRN~\cite{le2021hcrn} & - & 45.9 & 49.3 & 53.7 & 48.2 \\
        HGA~\cite{jiang2020hga} & -  & 46.3 & 50.7 & 59.3 & 49.7  \\
        VGT~\cite{xiao2022vgt} & 0.18M & 53.4 & 56.4 & 69.5 & 56.9 \\
        \demph{HGA*~\cite{jiang2020hga}} & \demph{400M} & \demph{46.8} & \demph{52.1} & \demph{59.3} & \demph{50.4} \\
        \demph{ATP~\cite{buch2022atp}} & \demph{400M} & \demph{51.3} & \demph{50.2} & \demph{66.8} & \demph{54.3}\\
        \hline
        \modelname & 5M & \textbf{62.4} & \textbf{58.3} & \textbf{75.6} & \textbf{63.1} \\
        
        \shline
        \textit{Hard Split} \\
        \hline
        \demph{ATP~\cite{buch2022atp}} & \demph{400M} & \demph{38.4} & \demph{36.5} & \demph{/} & \demph{/}\\
        \demph{HGA~\cite{jiang2020hga}} & \demph{-} & \demph{43.3} & \demph{45.3} & \demph{/} & \demph{/}\\
        \hline
        \modelname & 5M & \textbf{47.8} & \textbf{48.6} & \textbf{/} & \textbf{/} \\
        \shline
    \end{tabular}
    }
    \vspace{-1ex}
    \caption{Comparison of existing methods on NExT-QA~\cite{xiao2021nextqa}. We report accuracy on the Causal (C), Temporal (T), Descriptive (D) splits and overall accuracy on validation set. * stands for using CLIP as the initialization of visual encoder.}
    \label{table:videoqa-nextqa}
    \vspace{-2ex}
\end{table}

\subsection{Comparison to Prior Arts}
In this section, we compare \modelname with numerous state-of-the-art video-language pre-training methods on several downstream datasets under fine-tuning setting. 

\subsubsection{Text-to-Video Retrieval}
Table \ref{table:retrieval-sota} summarizes the results on MSRVTT \cite{xu2016msrvtt}, DiDeMo \cite{anne2017didemo}, LSMDC \cite{rohrbach2015lsmdc}, and ActivityNet Caption \cite{krishna2017activitynetret} under fine-tuning settings. Our method outperforms all of the existing video-language pre-training model by a large margin under the same data scale. In particular, our method yields 6.6\% lift in terms of R@1 on MSRVTT dataset while only exploiting 5M video-text pairs. Note that we also include the comparison with the recent works that utilize the powerful encoder from CLIP \cite{radford2021clip}, our method still can be comparable with them even surpass them, which shows the validness of the proposed method. Besides, we can notice that our method achieves the best result among all of listed methods on LSMDC dataset, which proves that our model can leverage the various moments presented in fruitful movie clips with cross-modal moment exploration. 

\subsubsection{Video Question Answering}
Table \ref{table:videoqa-sota} lists the results of \modelname and current state-of-the-art approaches on nine VideoQA datasets. It can be noticed that our method achieves the best performance in most of VideoQA datasets even with less pre-training data. Specifically, it achieves absolute improvement 1.1\% on TGIF-FrameQA, 2.2\% on MSRVTT-MC, 1.5\% on MSRVTT-QA, 0.2\% on MSVD-QA, and 3.3\% on ActivityNet-QA. We believe the moments learned by the cross-modal exploration are useful for finding the clue of answers in VideoQA.

\vspace{-2ex}
\subsubsection{Video Captioning}
\vspace{-1ex}
Table \ref{table:caption-sota} compares \modelname with existing mthods on video captioning datasets MSRVTT and MSVD.
As shown in the table, although we use less pre-training data than compared approaches, \modelname still obtains significant improvement compared to those large-scale pre-trained models. On MSRVTT Caption, our method surpasses SoTA method MV-GPT \cite{seo2022mvgpt} by 2.5\% CIDEr. Note that MV-GPT is pre-trained for multi-modal video captioning and it leverages the ASR transcripts from audio as the additional input. By contrast, our method only utilizes video as the input during generation.

\begin{table}[t!]
\centering
    \tablestyle{7pt}{1.1} 
    \def \w{15pt}
    \resizebox{0.9\linewidth}{!}{
        \begin{tabular}{lccc}
            \shline
            Dataset & \multicolumn{1}{c}{Original \color{blue}{$\uparrow$}} & \multicolumn{1}{c}{Shuffled \color{red}{$\downarrow$}} & \multicolumn{1}{c}{Gap \color{blue}{$\uparrow$}} \\
            \hline
            MSRVTT~\cite{xu2016msrvtt} & 64.2 & 63.3 & 0.9 \\
            DiDeMo~\cite{anne2017didemo} & 72.1 & 70.2 & 1.9 \\
            LSMDC~\cite{rohrbach2015lsmdc} & 42.6 & 41.7 & 0.9 \\
            ActivityNet Caption~\cite{krishna2017activitynetret} & 67.6 & 66.8 & 0.8 \\
            SSv2 Template~\cite{lei2022singularity} & 95.2 & 72.4 & 22.8 \\
            SSv2 Label~\cite{lei2022singularity} & 76.7 & 73.5 & 3.2 \\
            \shline
        \end{tabular}
    }
    \vspace{-1ex}
    \caption{Dependency on temporal information for text-to-video retrieval datasets with temporal shuffling test. The average recall of Recall@1, Recall@5, and Recall@10 are reported. We evaluate the performance drop when shuffling the input during inference. “Original” and “Shuffled” denote the original and shuffled input videos, respectively, and “Gap” is the difference between the Original and Shuffled metric. The larger "Gap" indicates the dataset relies on temporal information, and the model utilizes more temporal information to solve the task.}
    \label{table:retrieval-shuffled}
    \vspace{-5ex}
\end{table}
\begin{table}[t!]
\centering
    \tablestyle{7pt}{1.1} 
    \def \w{15pt}
    \resizebox{0.9\linewidth}{!}{
        \begin{tabular}{lccc}
            \shline
            Dataset & \multicolumn{1}{c}{Original \color{blue}{$\uparrow$}} & \multicolumn{1}{c}{Shuffled \color{red}{$\downarrow$}} & \multicolumn{1}{c}{Gap \color{blue}{$\uparrow$}} \\
            \hline
            MSRVTT-QA~\cite{xu2017msrvttqa} & 45.4 & 45.2 & 0.2 \\
            MSVD-QA~\cite{xu2017msrvttqa} & 55.6 & 55.5 & 0.1 \\
            TGIF-FrameQA~\cite{jang2017tgif} & 72.5 & 72.1 & 0.4 \\
            ActivityNet-QA~\cite{yu2019activitynetqa} & 45.1 & 45.0 & 0.1 \\
            NExT-QA (Hard)~\cite{xiao2021nextqa} & 47.1 & 45.6 & 0.5 \\
            \shline
        \end{tabular}
    }
    \vspace{-1ex}
    \caption{Dependency on temporal information for video question answering datasets by temporal shuffling test. We report the accuarcy for each dataset. For NExT-QA dataset, we evaluate with the hard split of the validation set~\cite{buch2022atp}.}
    \label{table:videoqa-shuffled}
    \vspace{-3ex}
\end{table}
\subsection{Discussion}
In this section, we discuss the temporal characteristics of our model and the datasets.
\paragraph{Impact of Loss Terms.} 
We investigate the contribution of individual loss terms and the results are shown in Table~\ref{table:ablation_loss}. It can be observed that the combining both $\mathcal{L}_{\text{CME}}$ and $\mathcal{L}_{\text{MTRE}}$ improves the performance of text-to-video retrieval and video question answering by at least 1.7\% and 2.9\% in Average Recall and Average accuracy respectively. In addition, we also find that the performance of $\mathcal{L}_{\text{CME}}$ surpasses that of $\mathcal{L}_{\text{MTRE}}$ on MSRVTT retrieval dataset that largely dominated by the appearance information. This can be explained that the cross-modal moment exploration loss not only select the positive verbs for the video from the text but also choose the acting object for alignment, which can boost the retrieval performance.

\paragraph{Evaluation on Temporal-aware Tasks.}
Lei \etal~\cite{lei2022singularity} reveal that the previous four retrieval datasets are prone to being biased for appearance while rarely relying on temporal information, thus introducing Something-to-Something v2 (SSv2) Template and SSv2 Label retrieval datasets to test models' true temporal modeling capability.
In particular, SSv2 Template retrieval task requires a deeper understanding of the moment and temporal relation since no objects information are presented. The performance on these datasets are summarized in Table \ref{table:retrieval-ssv2}. It can be observed that \modelname achieves significant improvement with $+8.5\%$ gains in terms of R@1 on these two temporal-oriented text-to-video retrieval datasets, which demonstrates the effectiveness of our proposed method through
exploring fine-grained moment information and modeling temporal relation. In addition, we evaluate our model on NExT-QA~\cite{xiao2021nextqa} dataset that explicitly designed for temporal and causal understanding. 
As presented in Table~\ref{table:videoqa-nextqa},
our method significantly surpasses its competitive counterparts, even those methods equipped with powerful image-text pre-trained encoders. Quantitatively, \modelname obtains an absolute improvement $+9\%$ on the causality split with the help of intrinsic temporal relation. Recently, Buch \etal~\cite{buch2022atp} filter out the trivial question for the dataset, and build the hard split for causality and temporal related questions for evaluate the causality and temporal of the model. As we can see in the table, even for the questions that heavily rely on causality, our model can still achieves a relative gain of 4.1\% on the model with specific design for VideoQA, which indicates that our model do not solely depend on static appearance.

\begin{figure*}
    \centering
    \includegraphics[width=1.0\textwidth]{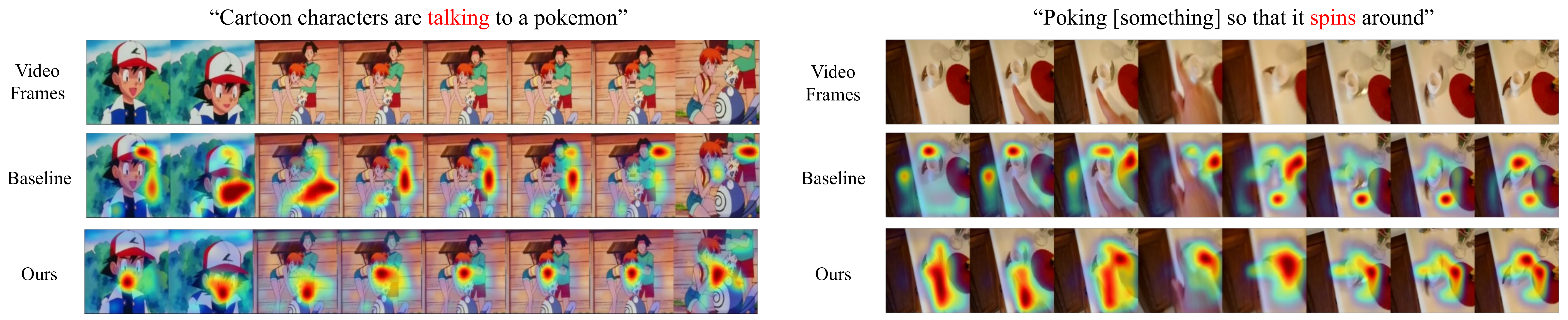}
    \vspace{-3ex}
    \caption{Visualizations of learned cross-attention maps from multi-modal encoder. We present samples from MSRVTT \cite{xu2016msrvtt} and SSv2 Template \cite{lei2022singularity} retrieval dataset. \modelname attends to the patches related to objects motion by tracking trajectory. Best viewed in color.}
\label{fig:ssv2_cam}
\vspace{-2ex}
\end{figure*}

\begin{table*}[t!]
\centering
    \tablestyle{7pt}{1.1} 
    \def \w{15pt}
    \resizebox{0.8\linewidth}{!}{
    \begin{tabular}{ll|ccc|ccc|ccc}
        \shline
        ~ & ~ & \multicolumn{3}{c}{MSRVTT}& \multicolumn{3}{c}{DiDeMo} & \multicolumn{3}{c}{LSMDC} \\
        \cmidrule(lr){3-5} \cmidrule(lr){6-8} \cmidrule(lr){9-11}
        Method & \# PT Data & R@1 & R@5 & R@10 & R@1 & R@5 & R@10 & R@1 & R@5 & R@10 \\
        \shline
        Frozen~\cite{bain2021frozen} & 5M  & 18.7 & 39.5 & 51.6 & 21.1 & 46.0 & 56.2 & 9.3 & 22.0 & 30.1 \\
        ALPRO~\cite{li2022alpro} & 5M & 24.1 & 44.7 & 55.4 & 23.8 & 47.3 & 57.9 & - & - & -  \\
        BridgeFormer~\cite{ge2022bridgeformer} & 5M  & 26.0 & 46.4 & 56.4 & 25.6 & 50.6 & 61.1 & 12.2 & 25.9 & 32.2 \\
        Singularity~\cite{lei2022singularity} & 5M & 28.4 & 50.2 & 59.5 & \textbf{36.9} & \textbf{61.6} & 69.3 & - & - & - \\
         \hline
         \multicolumn{5}{l}{\demph{\textit{Models pre-trained on more data}}}\\
         \hline
         \demph{VideoCLIP~\cite{xu2021videoclip}} & \demph{138M}  & \demph{10.4} & \demph{22.2} & \demph{30.0} & \demph{16.6} & \demph{46.9} & \demph{-}  & \demph{-} & \demph{-} & \demph{-} \\
        \demph{VIOLET~\cite{fu2021violet}} & \demph{183M}  & \demph{25.9} & \demph{49.5} & \demph{59.7} & \demph{23.5} & \demph{49.8} & \demph{59.8}  & \demph{-} & \demph{-} & \demph{-} \\
        \demph{Clip4Clip~\cite{luo2022clip4clip}} & \demph{400M} &  \demph{31.2} & \demph{53.7} & \demph{64.2} & \demph{-} & \demph{-} & \demph{-}  & \demph{11.3} & \demph{22.7} & \demph{29.2} \\
        \hline
        \modelname & 5M & \textbf{29.9} & \textbf{54.2} & \textbf{62.9} & 36.1 & 60.1 & \textbf{70.3} &  \textbf{15.5} & \textbf{31.1} & \textbf{39.8} \\
        \demph{\modelname} & \demph{17M} & \demph{\textbf{34.4}} & \demph{\textbf{60.0}} & \demph{\textbf{69.9}} & \demph{\textbf{43.2}} & \demph{\textbf{69.3}} & \demph{\textbf{79.0}} & \demph{\textbf{18.3}} & \demph{\textbf{36.7}} & \demph{\textbf{44.2}} \\
        \shline
    \end{tabular}
    }
    \vspace{-1ex}
    \caption{\textbf{Zero-shot evaluation on text-to-video retrieval.} All results are reported on R@1/R@5/R@10. We gray out methods that use significantly more pre-training data for fair comparison.
    }
    \label{table:retrieval-zeroshot}
    \vspace{-2ex}
\end{table*}


\begin{table}[t!]
\centering
    \tablestyle{7pt}{1.1} 
    \def \w{15pt}
    \resizebox{\linewidth}{!}{
    \begin{tabular}{ll|cc}
        \shline
        Method & \# PT Data & MSRVTT-QA & MSVD-QA \\
        \hline
        Just Ask~\cite{yang2021justask} & 69M & 2.9 & 7.5 \\
        LAVENDER~\cite{li2022lavender} & 5M & 4.5 & 11.6 \\
        MERLOT Reserve~\cite{zellers2022merlotreserve} & 1B & 5.8 & - \\
        FrozenBiLM~\cite{yang2022frozenbilm} & 10M & 6.4 & 11.7 \\
        \modelname & 5M & \textbf{8.6} & \textbf{18.2} \\
        \hline
        \demph{BLIP~}\cite{li2022blip} & \demph{129M}  & \demph{19.2} & \demph{35.2} \\
        \demph{mPLUG}~\cite{li2022mplug} & \demph{400M} & \demph{21.1}  & \demph{37.2} \\
        \demph{\modelname} & \demph{5M} & \demph{\textbf{21.7}} & \demph{\textbf{37.4}} \\
        \shline
    \end{tabular}
    }
    \vspace{-1ex}
    \caption{\textbf{Zero-shot evaluation on video question answering.} Accuracy is reported. We gray out those methods additionally supervised pre-training on VQA v2~\cite{goyal2017vqa} dataset.}
    \label{table:videoqa-zeroshot}
    \vspace{-3ex}
\end{table}

\paragraph{Temporal Reliance of Datasets.}
Previous methods~\cite{li2022alpro, lei2021clipbert, ge2022bridgeformer} only evaluate the performance of models on the existing datasets to demonstrate the superiority of the methods. However, Buch \etal~\cite{buch2022atp} and Lei \etal~\cite{lei2022singularity} reveal that the most of the evaluation are biased towards the static concepts. Here, we investigate the temporal reliance for the evaluated datasets by introducing the temporal shuffling test. Specifically, we compute the performance changes between running inference on the ordered video versus its shuffled version. The large performance drop indicates the dataset has less spatial bias and needs for temporal information. Table~\ref{table:retrieval-shuffled} and Table~\ref{table:videoqa-shuffled} conclude the performance gap between ordered and shuffled input video for text-to-video retrieval and VideoQA datasets. For text-to-video retrieval task, SSv2 Template shows the large performance drop after shuffling the input video, which demonstrates that it depends most on the dynamic information thus verifying our assumption. On the contrary, the performance on ActivityNet Caption dataset is barely affected (-0.8 on Mean Recall) since the text almost describes the static objects without relying on temporal information. For video question answering dataset, we observe that the MSVD-QA and ActivityNet-QA are less sensitive to the order of video frames. This is because these two datasets contain more questions requiring frame-region information, such as object categories, scenes, and species. We believe this can be used to evaluate the temporal reliance of the datasets as well as the utilization for temporal cue by models in the future work. 

\paragraph{Qualitative Analysis.}
To verify that our model can capture the motion information with respected to the given text rather than inferring from the static signal, we present the query text in SSv2 Template dataset which has masked all of the object information, and also visualize the query in MSRVTT dataset. As we can see in the Figure~\ref{fig:ssv2_cam}, the attention map of atomic action "talking" mainly focuses on the mouse of the cartoon characters while the baseline largely focusing on the characters, which indicates that our method can understand the moment better when adopting the temporal-aware pre-training tasks. In another example, the word "spins" can reveal the trajectory of the object showing that our method is able to capture the temporal motion presented in the video.

\vspace{-1ex}
\subsection{Zero-shot Generalizability}
\vspace{-1ex}
To demonstrate the generalizability of proposed video-text pre-trained model, we perform zero-shot evaluation on video-language downstream tasks. Table \ref{table:retrieval-zeroshot} summarizes the performance of our model and compared approaches on text-to-video retrieval. We can observe that our model yields more than 3.4\% lift in R@1 on MSRVTT dataset \cite{xu2016msrvtt} while exploiting less video-text pairs. Besides, our method surpasses all of the compared models in terms of LSMDC dataset showing the superiority of our model's generalizability. We also evaluate the zero-shot performance on VideoQA task in Table~\ref{table:videoqa-zeroshot}. Our method attains competitive zero-shot performance on MSRVTT-QA and MSVD-QA datasets even without help of audio signal supervision \cite{zellers2022merlotreserve} or additional generated video question pairs \cite{yang2021justask}. In particular, less pre-training data (\ie 5M $<$ 69M) are used while our method can still outperform other SoTA approaches. We also evaluate the zero-shot performance of models supervised on VQA v2 \cite{goyal2017vqa}. We can find that our method surpasses the powerful multi-modal SoTA methods (\eg mPLUG~\cite{li2022mplug}) with only 5M pre-training data showing the better generalization ability of \modelname.

\section{Conclusion}
In this work, we introduce \modelname, a novel hierarchical temporal-aware video-language pre-training framework with both understanding and generation capabilities. We vary the video with different views and model cross-modal alignment between moments and texts as well as their temporal relations in a hierarchical way.
Specifically, a \textit{cross-modal moment exploration} pre-training task is proposed to explore the alignment between the text and video moment, which helps to overcome the partially semantic alignment between video and text. Moreover, multi-modal pairs are constructed to learn temporal relations between moments and the event presented by the video with \textit{multi-modal temporal relation exploration} pre-training task. Even pre-trained on less data, \modelname still achieves state-of-the-art performance on a wide range of video-language downstream datasets, which clearly shows the superiority of our method.



{\small
\bibliographystyle{ieee_fullname}
\bibliography{egbib}
}

\clearpage
\appendix
\section{Additional Experimental Results}
In this section, we provide more experimental results for completeness of our proposed method.
\subsection{Transfer to Image-Text Downstream Tasks}
Since images can be viewed as the single-frame videos, we evaluate the proposed method on image-text tasks including image-text retrieval and visual question answering.
\paragraph{Image-Text Retrieval}
We perform the Image-to-Text and Text-to-Image retrieval on COCO datasets, and the results are summarized in Table \ref{tab:img_ret_coco}. We can observe that our method surpasses Singularity \cite{lei2022singularity} with same amount of pre-train data, especially 1\% improvement on Recall@1 for Text-to-Image retrieval task. Moreover, although some methods \cite{li2021align, chen2020uniter} leverage 4M dataset which contains the COCO dataset as a part of the pre-training dataset, \modelname can still attain comparable results showing the good generalization ability.

\paragraph{Visual Question Answering}
We also evaluate our method on visual question answering task. Table \ref{tab:img_vqa} concludes the image question answering results on VQAv2 \cite{goyal2017vqa} datasets. We observe that \modelname demonstrates competitive performance on the VQA tasks. It is worthwhile noting that our method achieves the better performance compared to Singularity \cite{lei2022singularity} same pre-training datasets, which indicates the video-text pre-training would boost the performance of image-text downstream tasks. However, we still see a gap with state-of-the-art image-text pre-trained models since our method do not use the in-domain data (\eg COCO) during pre-training, thus leading to the gap with SoTA performance. One future direction is to use more image-text data during video-text pre-training for better generalization.

\subsection{Additional Ablation Studies}

\begin{figure*}
    \centering
    \includegraphics[width=1.0\textwidth]{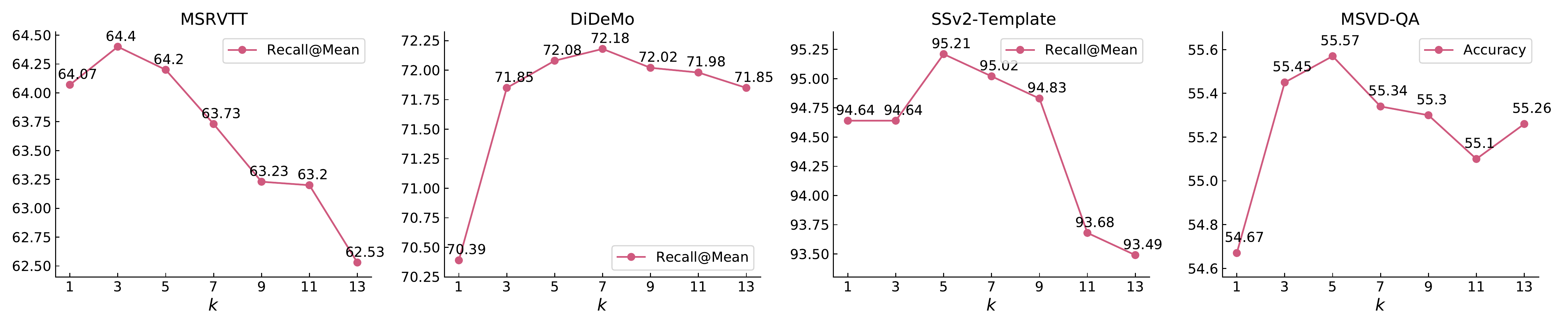}
    \caption{Variations in performance by changing the number of selected positive words $K$.}
\label{fig:topk}
\end{figure*}

\paragraph{Impact of positive candidate words size $K$.}
We investigate the effect of choosing different positive words size $K$ during cross-modal moment exploration. As depicted in Figure \ref{fig:topk}, it can be observed that with the increment of $K$, the performance on each dataset is increasing then start to decrease. In addition, there is a trade-off between the choice of $K$ and performance with respected to different datasets, and $K = 5$ gives relative good results among these datasets. It also suggests that the small $K$ would give more deterministic results since the model would only select the word with the largest similarity, thus more focusing on the single action or object. Then, as number of positive words increased, more accurate words are selected to align with the short-view of video. However, the model no longer benefits from cross-modal moment exploration when $K$ is large enough (\ie, $K = 11$ or $K = 13$) due to the increased noise in the selected candidate words.

\begin{table}[t]
\vspace{1em}
\tablestyle{3.5pt}{1.05}
\begin{tabular}{lrcccccc}
\shline
\multirow{3}{*}{ Method } & \multirow{3}{*}{ \#PT Data} & \multicolumn{6}{c}{COCO (5K test)} \\
& & \multicolumn{3}{c}{TR} & \multicolumn{3}{c}{IR } \\
\cmidrule(lr){3-5} \cmidrule(lr){6-8}
& & R1 & R5 & R10 & R1 & R5 & R10 \\
\midrule
ViLT~\cite{kim2021vilt} & 4M & 61.5 & 86.3 & 92.7 & 42.7 & 72.9 & 83.1 \\
UNITER~\cite{chen2020uniter} & 4M & 65.7 & 88.6 & 93.8 & 52.9 & 79.9 & 88.0 \\
OSCAR~\cite{li2020oscar} & 4M & 70.0 & 91.1 & 95.5 & 54.0 & 80.8 & 88.5 \\
ALBEF~\cite{li2021align} & 4M & 73.1 & 91.4 & 96.0 & 56.8 & 81.5 & 89.2 \\
BLIP~\cite{li2022blip} & 14M & 80.6 & 95.2 & 97.6 & 63.1 & 85.3 & 91.1 \\
ALIGN~\cite{jia2021scaling} & 1.2B & 77.0 & 93.5 & 96.9 & 59.9 & 83.3 & 89.8 \\
Singularity~\cite{lei2022singularity} & 5M & 71.9 & 90.8 & 95.4 & 54.6 & 80.0 & 87.8 \\
\hline
\modelname & 5M & 72.4 & 90.9 & 95.4 & 55.6 & 80.6 & 87.8 \\
\shline
\end{tabular}
\caption{Comparison to existing methods on image-text retrieval on COCO dataset. We show results for both text retrieval (image-to-text retrieval, TR) and image retrieval (IR).}
\label{tab:img_ret_coco}
\end{table}
\begin{table}[t]
\vspace{1em}
\tablestyle{7pt}{1.05}
\begin{tabular}{lrcc}
\shline
Method & \#PT Data & test-dev & test-std \\
\hline
ClipBERT~\cite{lei2021clipbert} & 0.2M & 69.08 & 69.43 \\
ViLT~\cite{kim2021vilt} & 4M & 70.94 & - \\
VL-BART~\cite{cho2021unifying} & 0.2M & - & 71.30 \\
LXMERT~\cite{tan2019lxmert} & 4M & 72.42 & 72.54 \\
UNITER~\cite{chen2020uniter} & 4M & 72.70 & 72.91 \\
UNIMO~\cite{li2020unimo} & 4M & 73.79 & 74.02 \\
OSCAR~\cite{li2020oscar} & 4M & 73.16 & 73.44 \\
ALBEF~\cite{li2021align} & 4M & 74.54 & 74.70 \\
BLIP~\cite{li2022blip} & 14M & 77.54 & 77.62 \\
Singularity~\cite{lei2022singularity} & 5M & 70.30 & 70.53 \\
\hline
\modelname & 5M & \textbf{74.06} & \textbf{74.28} \\
\shline
\end{tabular}
\caption{Comparison to existing methods on VQA.}
\label{tab:img_vqa}
\vspace{-2ex}
\end{table}

\begin{table*}[t!]
\centering
    \tablestyle{7pt}{1.1} 
    \def \w{15pt}
    \resizebox{0.95\linewidth}{!}{
        \begin{tabular}{lccccccccc}
            \shline
            
            ~ & \multicolumn{3}{c}{MSRVTT \cite{xu2016msrvtt}} & \multicolumn{3}{c}{SSv2-Template \cite{lei2022singularity}} & \multicolumn{3}{c}{SSv2-Label \cite{lei2022singularity}} \\
            \cmidrule(lr){2-4} \cmidrule(lr){5-7} \cmidrule(lr){8-10}

            Method & Original \color{blue}{$\uparrow$} & Shuffled \color{red}{$\downarrow$} & Gap \color{blue}{$\uparrow$} & Original \color{blue}{$\uparrow$} & Shuffled \color{red}{$\downarrow$} & Gap \color{blue}{$\uparrow$} & Original \color{blue}{$\uparrow$} & Shuffled \color{red}{$\downarrow$} & Gap \color{blue}{$\uparrow$} \\
            \hline
            $\mathcal{L}_{\text{base}}$ & 61.7 & 60.8 & 0.9 & 93.5 & 75.1 & 18.4 & 74.6 & 71.9 & 2.7\\
            $\mathcal{L}_{\text{base}}$ + $\mathcal{L}_{\text{CME}}$ & 63.7 & 62.9 & 0.8 & 94.4 & 72.6 & 21.8 & 74.8 & 71.8 & 3.0 \\
            $\mathcal{L}_{\text{base}}$ + $\mathcal{L}_{\text{MTRE}}$ & 63.0 & 62.6 & 0.4 & 94.1 & 73.0 & 21.1 & 75.8 & 72.2 & \textbf{3.6} \\
            $\mathcal{L}_{\text{base}}$ + $\mathcal{L}_{\text{CME}}$ + $\mathcal{L}_{\text{MTRE}}$ & 64.2 & 63.3 & \textbf{0.9} & 95.2 & 72.4 & \textbf{22.8} & 76.7 & 73.5 & 3.2 \\
            \shline
        \end{tabular}
    }
    \vspace{-1ex}
    \caption{Evaluation of proposed methods for temporal dependency with temporal shuffling test. We evaluate the performance drop when shuffling the input during inference. “Original” and “Shuffled” denote the original and shuffled input videos, respectively, and “Gap” is the difference between the Original and Shuffled metric. The larger "Gap" indicates the dataset relies on temporal information, and the model utilizes more temporal information to solve the task.}
    \label{table:ablation-shuffled}
    \vspace{-2ex}
\end{table*}

\paragraph{Temporal evaluation of loss terms.}
To further validate the temporal dependency for the proposed method, we adopt the shuffling test for models with different loss terms, as shown in Table \ref{table:ablation-shuffled}. Table \ref{table:ablation-shuffled} shows that our loss terms contribute more significantly when the dataset requires more temporal understanding. In concrete, $\mathcal{L}_{\text{CME}}$ and $\mathcal{L}_{\text{MTRE}}$ consistently improve the performances of Original and Gap on more temporal relied datasets (\ie SSv2-Template and SSv2-Label). For example, model with two loss terms largely surpasses the baseline model in the metric of Gap by achieving 4.4 and 0.5 improvement on SSv2-Template and SSv2-Label, respectively.


\begin{table}[t]
\centering
    \tablestyle{5pt}{1.1} 
    \def \w{20pt} 
    \resizebox{\linewidth}{!}{
    \begin{tabular}{l|ccc}
        \shline
        Method & MSRVTT & DiDeMo & SSv2-Template \\
        \hline
        TimeSformer ($\mathcal{L}_{\text{base}}$) & 57.30 & 62.38 & 92.91 \\ \hline
        + $\mathcal{L}_{\text{MTRE}}$ & 59.23 & 63.18 & 93.68 \\ \hline
        + $\mathcal{L}_{\text{CME}}$ & 59.03 & 63.78 & 93.30  \\ \hline
        + $\mathcal{L}_{\text{CME}}$ + $\mathcal{L}_{\text{MTRE}}$ & \textbf{59.93} & \textbf{65.34} & \textbf{94.25} \\ \hline
        \shline
    \end{tabular}
    }
    \caption{Effectiveness of the proposed methods on different video backbone. We use TimeSformer \cite{bertasius2021timesformer} pre-trained on ImageNet-21K \cite{ridnik2021imagenet} to verify the generalization ability of our proposed method. For text-to-video retrieval, the Mean Recall of Recall@1, Recall@5, and Recall@10 is reported. For video question answering task, we report the Top-1 accuracy.}
    \label{table:ablation_backbone}
\end{table}

\paragraph{Generalization to other vision backbone.}
Table \ref{table:ablation_backbone} shows that our proposed method is generalizable to different vision backbones. In details, we instantiate the video encoder with TimeSformer \cite{bertasius2021timesformer} pretrained on ImageNet-21K \cite{ridnik2021imagenet}. It can be observed both CME and MTRE consistently improve the model performance across the video backbones considered showing the generalization of proposed hierarchical temporal-aware pre-training framework. 
It is worth noting that, TimeSformer generates long video tokens compared to that of Multi-scale ViT \cite{li2022mvitv2}, which brings extra memory cost for the multi-modal encoder and decoder since the computation of self-attention is quadratic. This makes TimeSformer expensive to scale to more input frames with longer sequences.

\begin{figure}
    \centering
    \includegraphics[width=0.95\linewidth]{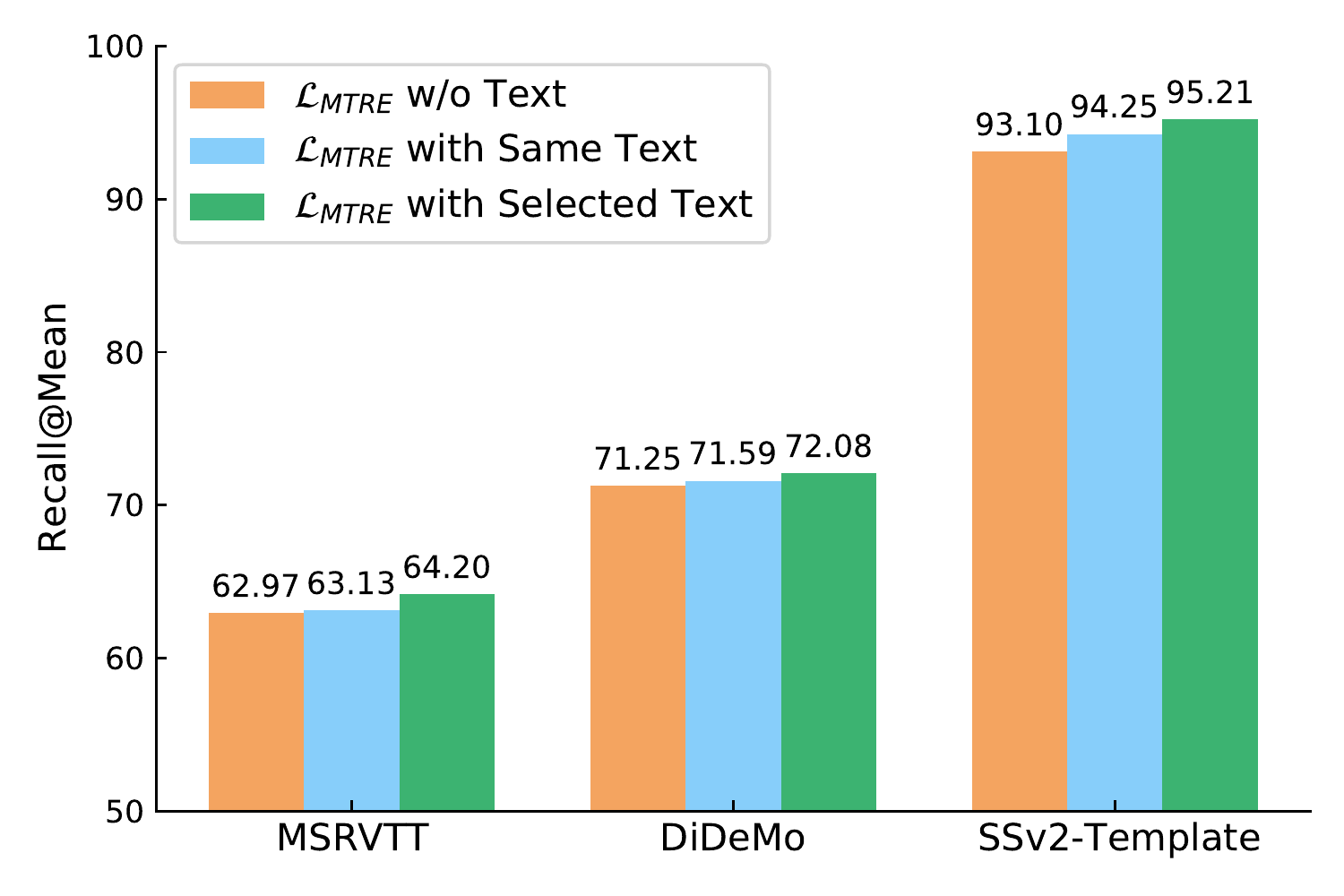}
    \caption{Variations in performance by adopting language during Multi-modal Temporal Relation Exploration (MTRE). We report the Mean Recall of Recall@1, Recall@5, and Recall@10.}
\label{fig:text_guidance}
\end{figure}

\paragraph{Influence of language for MTRE.}
We investigate the influence of language for multi-modal temporal relation exploration. Instead of utilizing the language signals, we directly adopt the video representation $v_{\text{cls}}$ from the video encoder during the learning. The results are sketched in Figure \ref{fig:text_guidance}. It can be observed that the model trained with multi-modal pairs attains better performance than the model without text. In concrete, it achieves 2.1\% gains on SSv2-Template which mainly depends on the understanding of actions, which indicates that our method can better understanding the actions via multi-modal temporal relation exploration. Besides, we also notice that performance of the model trained with correct multi-modal pairs surpasses that of model trained by multi-modal pairs with same text, which indicates that improper video-text pair yields noisy multi-modal representation thus degrading the performance of the model.

\begin{figure*}
    \centering
    \includegraphics[width=1\linewidth]{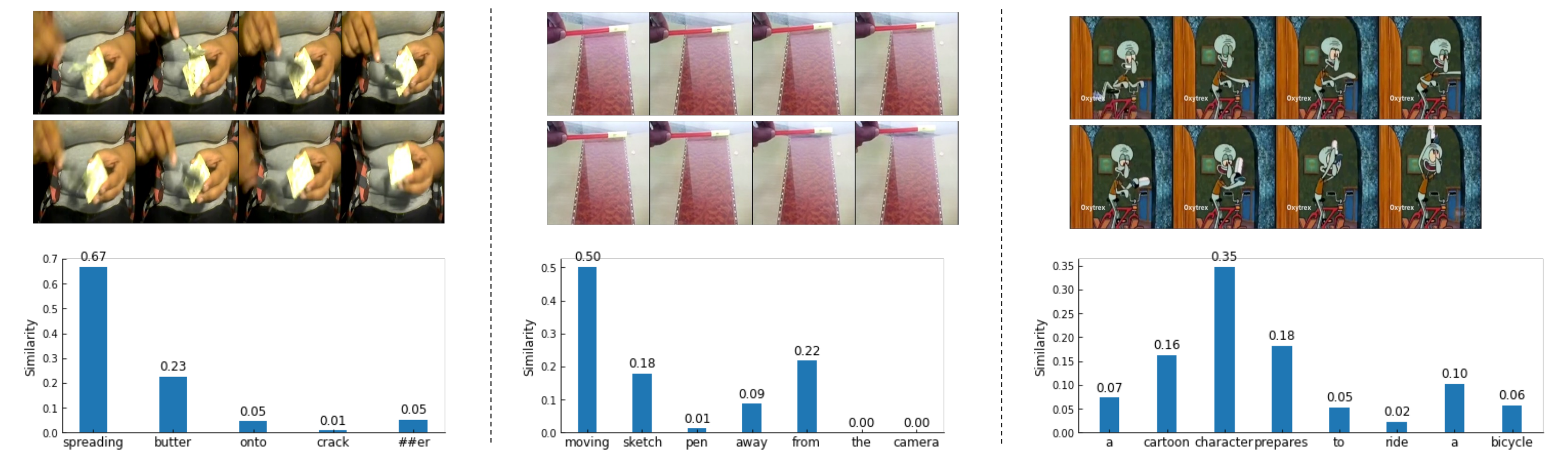}
    \vspace{-1ex}
    \caption{Examples of similarities between words and videos generated by our method. Our method captures the atomic actions in the videos as well as the object information with the help of cross-modal moment exploration.}
\label{fig:topk_words}
\end{figure*}

\section{Discussion}
\subsection{Qualitative Analysis}
We sample some videos and corresponding texts and compute similarities between words and videos in Figure \ref{fig:topk_words}. As we can see in the figure, our model can effectively capture the moments such as "spreading", "moving", and "preparing" etc. in the video, which is essential for understanding videos. Besides, we can notice that the video would also attend the object appeared in the video, showing the capability for modeling fine-grained moment information.

\subsection{Connection to Other Fine-Grained Methods}
Some efforts \cite{lee2018scan, yao2021filip, messina2021teran} have been made to learn the fine-grained correlation and alignment between two modalities by leveraging the token-wise similarities in vision-language pre-training. FILIP \cite{yao2021filip} and TERAN \cite{messina2021teran} aggregate the maximum token similarity scores and assign the optimal patch-word transport matrix. SCAN \cite{lee2018scan} utilizes the similarity scores to attend each tokens for soft fine-grained alignment. These approaches are originally tailored for image-text pre-training, which aims to locate the fine-grained static object. However, different from image-text pre-training, video-text pre-training needs to understand the correlation between words and moments, which not only contains static objects but also consists of atomic actions. Our proposed cross-modal moment exploration leverages the short-view of video to reflect the moment information and discover the relationship between short-view videos and words, which results in fine-grained moment representations for video-language pre-training.

\subsection{Limitations and Boarder Impact}
Despite the effectiveness of the proposed method on various downstream tasks, our method still has some limitations that would make for promising directions for future work. (1) Currently, we only pre-train our model on 5M data with the base-size encoders, and the scalability of the model is not explored which deserves more in-depth investigation in the future. (2) Our method shares similar risks like other pre-training methods that the pre-training data might consist bias and unsafe content which requires further analysis before the deployment.

\section{Implementation Details}
\subsection{Number of Parameters}
\begin{table}[t]
\vspace{1em}
\tablestyle{7pt}{1.05}
\begin{tabular}{lc}
\shline
Method & \# of Parameter \\
\hline
ClipBERT~\cite{lei2021clipbert} & 137M \\
Frozen~\cite{bain2021frozen} & 232M \\
BridgeFormer~\cite{ge2022bridgeformer} & 152M \\
All-in-one~\cite{wang2022allinone} & 110M \\
VIOLET~\cite{fu2021violet} & 198M \\
ALPRO~\cite{li2022alpro} & 231M \\
Singularity~\cite{lei2022singularity} & 209M \\
LAVENDER~\cite{fu2021violet} & 198M \\
\hline
\modelname & 297M \\
\shline
\end{tabular}
\caption{Comparison to other models in the number of parameters.}
\label{tab:model_params}
\vspace{-2ex}
\end{table}

We include some of previous models with their parameter counts (which were reported in the original paper or calculated by follow-up work), and we compare them with \modelname in Table \ref{tab:model_params}. Compared with other models, our model is of comparable model size and requires less pair of video-text pre-training data to achieve better performance in terms of both video-language understanding and generation.

\begin{figure}
    \centering
    \includegraphics[width=0.95\linewidth]{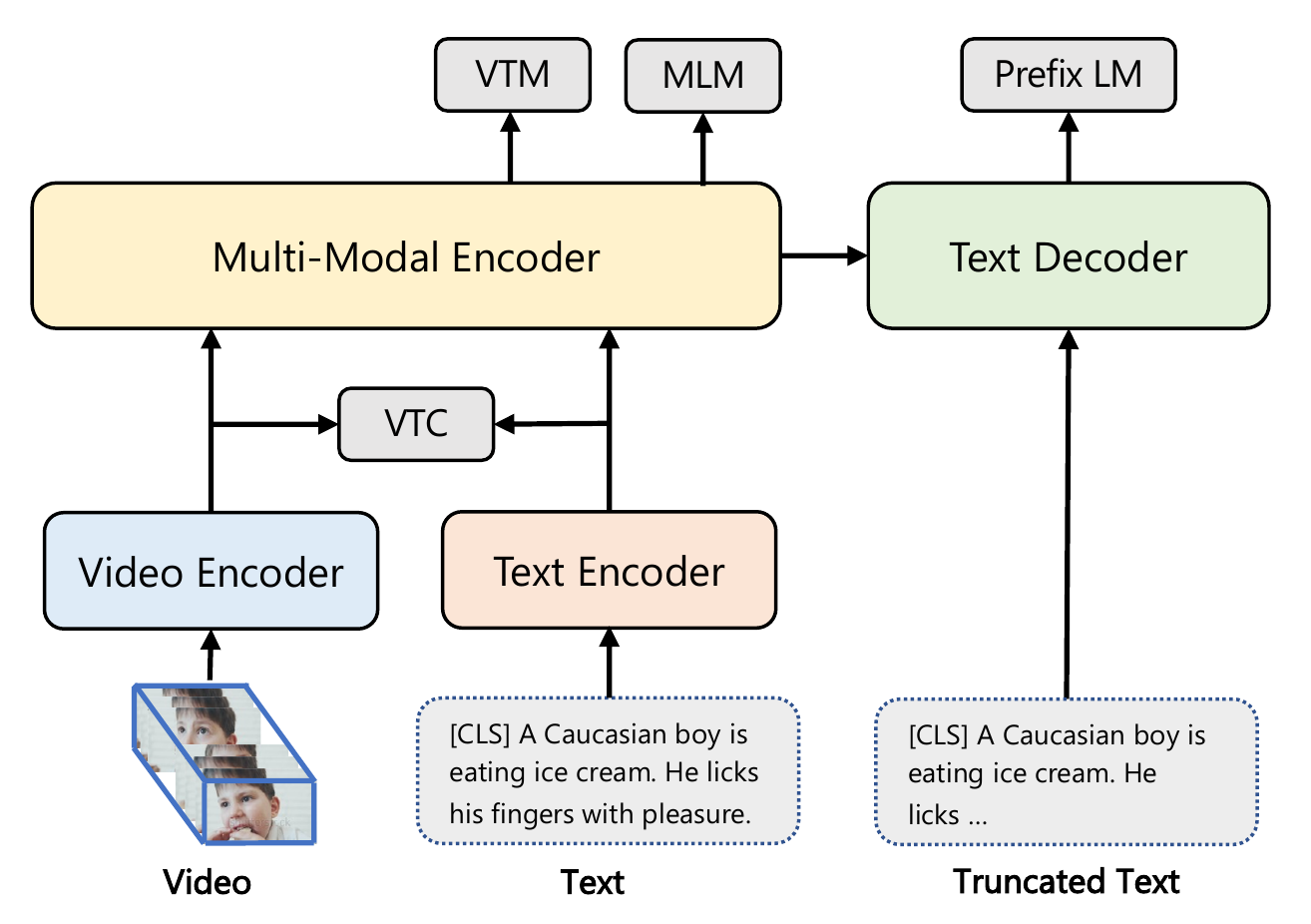}
    \caption{Architecture of the proposed \modelname and other pre-training objectives.}
\label{fig:model}
\end{figure}

\subsection{Model Architecture}
As sketched in Figure \ref{fig:model}, our model consists two unimodal encoders for text and video respectively, a multi-modal encoder for video-text interaction, and a text decoder for generation. In concrete, given an arbitrary view of video $V \in \mathbb{R}^{T\times H \times W}$ is encoder into a sequence of embeddings: $\{v_{\text{cls}}, v_1, \cdots, v_M\} \in \mathbb{R}^{(M+1) \times D}$, where $M$ is the number of flattened patches for video $V$, and $v_{\text{cls}}$ is the embedding of the visual [CLS] token and used to provide global representation of the video. The text encoder transforms the text into a sequence of embeddings: $\{w_{\text{cls}}, w_1, \cdots, w_N\} \in \mathbb{R}^{(N+1)\times D}$, where $N$ is the number of words in the text. To efficiently encode multi-modal information while preserving unimodal information, we fuse the video and text features from uni-modal encoders following~\cite{li2022mplug}. The output of the multi-modal encoder $\{\bm{v}_{\text{cls}}, \bm{v}_1, \cdots, \bm{v}_M, \bm{w}_{\text{cls}}, \bm{w}_1, \cdots, \bm{w}_N\} \in \mathbb{R}^{(M+N+2) \times D}$ is fed into a transformer decoder for sequence to sequence generation, which equips \modelname with the capabilities of both multi-modal understanding and generation. 

\subsection{Pre-training Objectives}
During pre-training, we also perform four pre-training tasks including Video-Text Contrastive Learning ($\mathcal{L}_{\text{VTC}}$), Video-Text Matching ($\mathcal{L}_{\text{VTM}}$), Masked Language Modeling ($\mathcal{L}_{\text{MLM}}$), and Prefix Language Modeling ($\mathcal{L}_{\text{PrefixLM}}$). The VTC task first is applied to align the unimodal representation of video and text. And the multi-modal representation can be learned by VTM and MLM tasks. Upon on the video-language representations obtained from multi-modal encoder, the decoder is trained by PrefixLM loss with text completion task.
\paragraph{Video-Text Contrast (VTC)}
Following \cite{li2022alpro, wang2022allinone}, we align the unimodal encoders via this task. Specially, the softmax-normalized video-to-text and text-to-video similarities are computed, and we employ memory queues in MoCo \cite{chen2021moco} to increase the number of negative samples during learning. 
Formally, the video-text contrastive loss is calculated as:
\begin{align}
    \mathcal{L}_{v2t} &= -\frac{1}{B}\sum_{i=1}^B \log \frac{\exp (s(\mathcal{V}_i, \mathcal{T}_i))}{\sum_{j=1}^B \exp (s(\mathcal{V}_i, \mathcal{T}_j))}, \\ \nonumber
    \mathcal{L}_{t2v} &= -\frac{1}{B}\sum_{i=1}^B \log \frac{\exp (s(\mathcal{V}_i, \mathcal{T}_i))}{\sum_{j=1}^B \exp (s(\mathcal{V}_j, \mathcal{T}_i))}, \\ \nonumber
    \mathcal{L}_{\text{VTC}} &= \frac{1}{2} (\mathcal{L}_{v2t} +\mathcal{L}_{t2v}),
\end{align}
where $\mathcal{V}_i$ and $\mathcal{T}_j$ are the projected representations of $v_{\text{cls}}$ and $w_{\text{cls}}$ for $i$-th video-text pair in the batch.

\paragraph{Video-Text Matching (VTM)}
This task aims to predict whether a video and a text is paired or not based on the multi-modal representation. As suggested in \cite{li2021align, li2022alpro}, hard negative video-text pairs are selected based on the similarity of video and text during contrastive learning.
Formally, the video-text matching loss is calculated as:
\begin{equation}
    \mathcal{L}_{\text{VTM}} = -\mathbb{E}_{(\mathcal{W}, \mathcal{V})} \log p(y | \mathcal{W}, \mathcal{V}),
\end{equation}
where $\mathcal{W}$ denotes the word tokens, and $\mathcal{V}$ denotes the video features of long-view video.

\paragraph{Masked Language Modeling (MLM)}
The setup of this pre-training task is same as that used in BERT \cite{devlin2018bert}, where 15\% of tokens in the text are randomly masked, and the model needs to predict the masked tokens based on the multi-modal representation.
Formally, the masked language modeling loss is calculated as:
\begin{equation}
    \mathcal{L}_{\text{MLM}} = -\mathbb{E}_{(\mathcal{W}, \mathcal{V})} \log p(w_i | \mathcal{W}_{\backslash i}, \mathcal{V}),
\end{equation}
where $w_i$ denotes the masked word token.

\paragraph{Prefix Language Modeling (PrefixLM)} This pretext task requires model to complete the truncated texts based on given videos and prefix sequence of truncated texts \cite{li2022mplug, li2022blip}. The model can be trained by maximizing the likelihood of the truncated text in an auto-regressive manner. 
Formally, the prefix language modeling loss is calculated as:
\begin{equation}
    \mathcal{L}_{\text{PrefixLM}} = -\mathbb{E}_{(\mathcal{W}, \mathcal{V})} \left[ \sum_{l=L_p}^L \log p(w_l | \mathcal{W}_{[L_p, l)}, \mathcal{W}_{<L_p}, \mathcal{V}) \right],
\end{equation}
where $L$ denotes the total number of words in the text, and $L_p$ is the length of a prefix sequence of tokens which is randomly selected.

\begin{table*}[t]
\vspace{1em}
\tablestyle{7pt}{1.05}
\begin{tabular}{l|cccccc}
\shline
\textbf{Dataset} & \textbf{Optimizer} & \textbf{Learning Rate} & \textbf{Weight Decay} & \textbf{LR Schedule} & \textbf{Batch Size $\times$ \# GPUs} & \textbf{Epochs}\\
\hline
MSRVTT-Ret \cite{xu2016msrvtt} & AdamW & 2e-5 & 0.02 & Cosine Decay & $24 \times 8$ & 10 \\
DiDeMo \cite{anne2017didemo} & AdamW & 1e-5 & 0.02 & Cosine Decay & $24 \times 8$ & 20 \\
LSMDC \cite{rohrbach2015lsmdc} & AdamW & 2e-5 & 0.02 & Cosine Decay & $24 \times 8$ & 10 \\
Activity Caption \cite{krishna2017activitynetret} & AdamW & 2e-5 & 0.02 & Cosine Decay & $24 \times 8$ & 20 \\
SSv2-Template \cite{lei2022singularity} & AdamW & 5e-5 & 0.02 & Cosine Decay & $24 \times 8$ & 20 \\
SSv2-Label \cite{lei2022singularity} & AdamW & 2e-5 & 0.02 & Cosine Decay & $24 \times 8$ & 20 \\

MSRVTT-QA \cite{xu2017msrvttqa} & AdamW & 2e-5 & 0.02 & Cosine Decay & $16 \times 8$ & 8 \\
MSVD-QA \cite{xu2017msrvttqa} & AdamW & 2e-5 & 0.02 & Cosine Decay & $16 \times 8$ & 8 \\
TGIF-FrameQA \cite{jang2017tgif} & AdamW & 2e-5 & 0.02 & Cosine Decay & $16 \times 8$ & 8 \\
LSMDC-FIB \cite{maharaj2017lsmdcfib} & AdamW & 2e-5 & 0.02 & Cosine Decay & $16 \times 8$ & 8 \\
ActivityNet-QA \cite{yu2019activitynetqa} & AdamW & 2e-5 & 0.02 & Cosine Decay & $16 \times 8$ & 8 \\

TGIF-Action \cite{jang2017tgif} & AdamW & 3e-5 & 0.02 & Cosine Decay & $16 \times 8$ & 56 \\
TGIF-Transition \cite{jang2017tgif} & AdamW & 3e-5 & 0.02 & Cosine Decay & $16 \times 8$ & 30 \\
LSMDC-MC \cite{torabi2016lsmdcmc} & AdamW & 2e-5 & 0.02 & Cosine Decay & $16 \times 8$ & 10 \\
NExT-QA \cite{xiao2021nextqa} & AdamW & 2e-5 & 0.02 & Cosine Decay & $16 \times 8$ & 10 \\

MSRVTT-Caption \cite{xu2016msrvtt} & AdamW & 2e-5 & 0.02 & Cosine Decay & $24 \times 8$ & 10 \\
MSVD-Caption \cite{chen2011msvd} & AdamW & 2e-5 & 0.02 & Cosine Decay & $24 \times 8$ & 10 \\
\shline
\end{tabular}
\caption{End-to-end fine-tuning configurations for video-language downstream tasks.}
\label{tab:finetune_params}
\end{table*}

\subsection{Downstream Task Implementation Details}
We evaluate \modelname on various downstream video-language tasks, including Text-to-Video Retrieval, Open-ended VideoQA, Multiple Choice VideoQA, and Video Captioning. The fine-tuning procedures are described as follows:

\begin{itemize}\setlength{\itemsep}{-1pt}
    \item For retrieval tasks, we jointly optimize the VTC loss and VTM loss for video-text alignment during fine-tuning. During inference, we first select top-k candidates by computing the dot-product similarity between the video and text features, and then reranking the selected candidates based on their VTM scores. $k$ is set to 128 by default.
    \item For open-ended VideoQA, we first generate video features and text features with two unimodal encoders, and then fuse them with multi-modal encoder. The output of multi-modal features are fed to text decoder for answer generation. We use the language modeling loss to optimize the model. During inference, the answer would be generated by the text decoder.
    \item For multiple choice VideoQA, we treat the problem as the text-to-video retrieval task where the correct answer should have the highest matching probability. During training, we compute the VTM scores for each candidate answer and video, then optimize the model with cross entropy loss. During the inference, the answer with highest VTM score is the prediction answer.
    \item For Video Captioning, we use the video features from video encoder and directly feed it into text decoder for caption generation. The language modeling loss is utilized for model optimization.
\end{itemize}

For all above video-language downstream tasks, we resize video frames to $224 \times 224$. During fine-tuning, following \cite{lei2022singularity, li2022alpro}, we randomly sample 12 frames for text-to-video retrieval, 16 frames for video question answering and video captions. We perform uniform sampling during inference. We use RandomCrop with minimum ratio 0.5 and HorizontalFlip with 0.5 probability for data augmentation. The hyperparameters that we used for fine-tuning on downstream tasks are summarized in Table \ref{tab:finetune_params}. For the video caption task, we use a prefix prompt “A video of” to improve the quality of generated captions.

\subsection{Datasets Description}
In this section, we describe all of the downstream video-language datasets used during evaluation. The details of the datasets are represented below:
\paragraph{Text-to-Video Retrieval.}
We evaluate \modelname on 6 popular text-to-video retrieval datasets including MSRVTT \cite{xu2016msrvtt}, DiDeMo \cite{anne2017didemo}, LSMDC \cite{rohrbach2015lsmdc}, ActivityNet Caption \cite{krishna2017activitynetret}, SSv2 Template \cite{lei2022singularity}, and SSv2 Label \cite{lei2022singularity}. Details of these datasets:
\textbf{MSRVTT} \cite{xu2016msrvtt} contains 10K YouTube sourced videos with 200K text descriptions. Following \cite{li2022lavender, luo2022clip4clip, huang2022clover}, we train the video on 9K videos and evaluate on the rest 1K video. 
\textbf{DiDeMo} \cite{anne2017didemo} contains of 10K videos from Flickr and 4 descriptions for each video. Following \cite{li2022alpro, ma2022xclip, li2022lavender}, we concatenate all of the given descriptions from the same video as a paragraph, and evaluate the paragraph-to-video retrieval performance. The number of video in training set is 8K, leaving 1K for validation set and 1K for test set.
\textbf{LSMDC} \cite{rohrbach2015lsmdc} consists of 118K video clips from 202 movies, and each clip is accompanied with a caption from video scripts. It has 101K video clips for training and 1K clips for testing. We use the standard splits from \cite{rohrbach2015lsmdc}.
\textbf{ActivityNet Caption} \cite{krishna2017activitynetret} is built on 20K YouTube videos with 100K captions. We use the train split with 10K videos for training, and report the performance on the val1 split with 4.9K videos. 
\textbf{SSv2-Template} and \textbf{SSv2-Label} \cite{lei2022singularity} contain 169K videos for training and 2K videos for testing. The text queries in SSv2-Template are templates without object information (\eg "Throwing [something] in the air and catching it"). By contrast, SSv2-Label contains annotated text queries with specific object information (\eg. "Throwing keys in the air and catching it"). Therefore, SSv2-Template mainly focuses on temporal understanding of actions, while SSv2-Label needs a more comprehensive understanding of both appearance and temporal dynamic.

\paragraph{Multiple-choice Video QA.} Five datasets are evaluated for multiple-choice video question answering tasks.
\textbf{TGIF-Action} and \textbf{TGIF-Transition} \cite{jang2017tgif} are adopted to evaluate model's capability to recognize the repeated actions and state transitions in short GIFs. Each video and question is equipped with 5 candidate answers. We concatenate the question and answer as the text and use the highest similarity among the video and candidate texts. TGIF-Action contains 18K GIFs for training and 2K for testing. TGIF-Transitions has 47K GIF-question pairs for training and 6K for testing.
\textbf{MSRVTT-MC} \cite{yu2018msrvttmc} and \textbf{LSMDC-MC} \cite{torabi2016lsmdcmc} are originally retrieval task, but reformulated as the multiple choice video QA task. It requires the model to find the optimal caption that describes the video out of 5 candidate texts. 
\textbf{NExT-QA} \cite{xiao2021nextqa} is explicitly designed for temporal
and causal understanding. Questions in the dataset are categorized into three types: Descriptive, Temporal, and Causal. Each question in the dataset are paired with 5 candidate answers. Therefore, this dataset is able to evaluate model's ability in video question answering in different aspects.

\paragraph{Open-ended Video QA.} For open-ended video QA, we evaluate the model on five datasets. 
\textbf{MSRVTT-QA} is composed of 243K open-ended questions over 10K videos, while \textbf{MSVD-QA} \cite{xu2017msrvttqa} consists 2K videos with 47K questions. \textbf{TGIF-Frames} \cite{jang2017tgif} collects the answerable with just a single frame in the video, and is divided into training set with 35K questions and test set with 14K questions.. For LSMDC-FiB \cite{maharaj2017lsmdcfib}, the model needs to predict a correct word for the blank with a given video and a sentence with blank. It contains 297K sentences for training and 30K sentences for testing. \textbf{ActivityNet-QA} \cite{yu2019activitynetqa} .

\paragraph{Video Captioning.} We use MSRVTT \cite{xu2016msrvtt} and MSVD \cite{chen2011msvd} for video captioning evaluation. As described before, \textbf{MSRVTT} is composed of 10K videos with 20 captions per video, and \textbf{MSVD} contains 2K videos with around 40 captions per video. We follow the standard splits from \cite{lin2022swinbert, li2022lavender}. During inference, we generate the caption with beam search until the model outputs a [SEP] that indicates the end of sentence or when it reaches the maximum generation step 40.

\end{document}